\documentclass[a4paper,fleqn]{cas-dc}

\usepackage{graphicx}
\usepackage[font=scriptsize, justification=centering]{caption}
\usepackage{subcaption}
\usepackage[linesnumbered, ruled,vlined]{algorithm2e}
\usepackage[switch]{lineno}
\usepackage{multirow}
\usepackage{enumitem}
\usepackage{soul}
\usepackage{graphicx}
\usepackage{subcaption}
\usepackage{setspace}
\usepackage{listings}
\usepackage{xspace}
\usepackage{tcolorbox}
\usepackage{mdframed}
\usepackage[flushleft]{threeparttable}
\usepackage[numbers]{natbib}
\usepackage{xcolor}

\def\tsc#1{\csdef{#1}{\textsc{\lowercase{#1}}\xspace}}
\tsc{WGM}
\tsc{QE}
\tsc{EP}
\tsc{PMS}
\tsc{BEC}
\tsc{DE}

\newdefinition{definition}{Definition}

\newcommand{\tool}{\textsc{Expona}\xspace}

\newcommand{\alchemist}{\textsc{Alchemist}\xspace}
\newcommand{\labelprop}{\textsc{Label Prop.}\xspace}
\newcommand{\fewshot}{\textsc{Few-shot Learn.}\xspace}
\newcommand{\datasculpt}{\textsc{DataSculpt}\xspace}
\newcommand{\snuba}{\textsc{Snuba}\xspace}

\newcommand{\mv}{\textsc{MultiVote}\xspace}
\newcommand{\wmv}{\textsc{W. MultiVote}\xspace}
\newcommand{\ds}{\textsc{DawidSkene}\xspace}
\newcommand{\fs}{\textsc{FlyingSquid}\xspace}
\newcommand{\snorkel}{\textsc{Snorkel}\xspace}

\definecolor{light-gray}{gray}{0.92} 
{\begin{mdframed}[backgroundcolor=light-gray,
skipabove=5pt,
skipbelow=0pt,
nobreak=false
]\begin{mdtheorem}{name}{label}}%
{\end{mdtheorem}\end{mdframed}}
\definecolor{ao}{rgb}{0.0, 0.5, 0.0}
\begin{document}
\let\WriteBookmarks\relax
\def\floatpagepagefraction{1}
\def\textpagefraction{.001}

\shorttitle{\tool}

\shortauthors{ \textit{Lam et~al.}}

\title [mode = title]{Structured Exploration and Exploitation of Label Functions for Automated Data Annotation}

\author{Phong Lam}
[orcid=0009-0002-5745-3361]
\ead{22028164@vnu.edu.vn}
\affiliation{organization={Faculty of Information Technology, VNU University of Engineering and Technology},
    city={Hanoi},
    country={Vietnam}}

\author{Ha-Linh Nguyen}
[orcid=0009-0007-3748-4810]
\ead{22024505@vnu.edu.vn}

\author{Thu-Trang Nguyen}
[orcid=0000-0002-3596-2352]
\ead{trang.nguyen@vnu.edu.vn}

\author{Son Nguyen}
[orcid=0000-0002-8970-9870]
\ead{sonnguyen@vnu.edu.vn}\cormark[1]

\author{Hieu Dinh Vo}
[orcid=0000-0002-9407-1971]
\ead{hieuvd@vnu.edu.vn}

\cortext[cor1]{Corresponding author}

\begin{abstract}
High-quality labeled data is critical for training reliable machine learning and deep learning models, yet manual annotation remains costly and error-prone. Programmatic labeling addresses this challenge by using label functions (LFs), i.e., heuristic rules that automatically generate weak labels for training datasets. However, existing automated LF generation methods either rely on large language models (LLMs) to synthesize surface-level heuristics or employ model-based synthesis over hand-crafted primitives. These approaches often result in limited coverage and unreliable label quality.  

In this paper, we introduce \tool, an automated framework for programmatic labeling that formulates LF generation as a principled process balancing \textit{diversity} and \textit{reliability}. \tool systematically explores multi-level LFs, spanning surface, structural, and semantic perspectives. \tool further applies reliability-aware mechanisms to suppress noisy or redundant heuristics while preserving complementary signals.

To evaluate \tool, we conducted extensive experiments on eleven classification datasets across diverse domains. Experimental results show that \tool consistently outperformed state-of-the-art automated LF generation methods. Specifically, \tool achieved nearly complete label coverage (up to 98.9\%), improved weak label quality by up to 87\%, and yielded downstream performance gains of up to 46\% in weighted F1.

These results indicate that \mbox{\tool's} combination of multi-level LF exploration and reliability-aware filtering enabled more consistent label quality and downstream performance across diverse tasks by balancing coverage and precision in the generated LF set.

\end{abstract}


\maketitle

\section{Introduction}

Machine learning (ML) and deep learning (DL) systems rely fundamentally on the availability of large-scale, high-quality labeled datasets~\cite{data-centric-ai,data-quality}. 
However, manual data labeling is notoriously expensive, time-consuming, and often error-prone, particularly in domains where expert knowledge is required or label semantics are subtle and context-dependent~\cite{gpt-3-labeling,annollm}. 
%
%
Consequently, there has been growing interest in \textit{automated data annotation}, which aims to reduce or eliminate human labeling effort through techniques such as active learning~\cite{active-learning-book, active-learning-survey, active-learning-survey-2, guan2024activedp, al-hybridization, tbal}, semi-supervised learning~\cite{wang2006label, semi, zhou2003learning}, programmatic labeling~\cite{data-programming, fu2020fast, goggles, interactive-ws, ratner2017snorkel, ruan2025, zhang2024stronger}, and LLM-as-annotator~\cite{gpt-3-labeling, annollm, LLM-few-short, zerogen, tuning-lm, llmaaa, schroeder2025}.

Among these, \textit{programmatic labeling}~\cite{data-programming, fu2020fast, ratner2017snorkel, mazuelas2025reliable} has emerged as a powerful paradigm for constructing weakly labeled datasets at scale. 
Instead of annotating individual samples, domain experts write a set of \textit{Label Functions (LFs)}, which are heuristics, patterns, or rules that assign noisy/weak labels or abstain on uncertain cases. 
A probabilistic label model then estimates the underlying true labels by reconciling conflicts among multiple LFs. This approach significantly reduces human workload by shifting the annotation process from instance-level labeling to heuristic design.
Nevertheless, the quality of the resulting dataset is highly dependent on the coverage, accuracy, and diversity of the LFs. Designing effective LFs requires substantial domain expertise and manual effort. Suboptimal or redundant LFs can severely degrade the label model's accuracy.

Recent studies have explored \textit{Automated LF Generation}, including model-based synthesis approaches~\cite{varma2018snuba, zhao2021glara} or Large Language Models (LLMs)-based methods~\cite{huang2024alchemist, guan2025datasculpt, li2025refining, smith2024language, alvarez2025automated} to construct labeling heuristics. While these methods represent important progress toward reducing manual effort in weak supervision, several key challenges remain.
Model-driven or graph-based methods~\cite{varma2018snuba, zhao2021glara} typically rely on predefined primitives, feature boundaries, or fixed search space. Such assumptions restrict their generality and reduce adaptability across datasets and tasks with different semantic structures. As a result, these methods often fail to capture complex or task-specific labeling patterns beyond predefined representations.
Recent LLM-based approaches such as \mbox{\alchemist~\cite{huang2024alchemist}} and \mbox{\datasculpt~\cite{guan2025datasculpt}} improve flexibility by generating labeling heuristics directly from natural language descriptions or data samples. However, the resulting heuristics are often shallow, surface-level, relying on lexical or local patterns while overlooking deeper structural or semantic relationships required for complex tasks. This can lead to brittle or overly specific labeling functions that generalize poorly.

Furthermore, existing approaches typically emphasize generation quantity or heuristic coverage without a principled mechanism for \textit{reliability assessment}. Consequently, noisy, redundant, or conflicting LFs are frequently introduced,  which negatively affects downstream label aggregation and reduces the overall effectiveness of weak supervision.
In essence, current automated LF generation methods struggle to jointly ensure the diversity and reliability of generated LFs, often producing heuristics that are either overly narrow or insufficiently trustworthy.
%

To address these limitations, we propose \tool, an automated framework for programmatic labeling that formulates LF generation as a principled process balancing \textit{diversity} and \textit{reliability}. 
Our key insight is that effective weak supervision requires a \textit{diverse yet reliable ensemble} of LFs, an aspect that has not been jointly addressed by prior automated LF generation approaches. Diversity ensures that the framework captures complementary task signals across \textit{surface}-, \textit{structural}-, and \textit{semantic}-level, thereby improving coverage. 
Meanwhile, reliability ensures that the LF set does not contain noisy or low-quality heuristics, preventing error propagation during label aggregation.

To implement this idea,  \mbox{\tool} adopts a two-phase process of \textit{LF exploration} and \textit{LF exploitation}. In the first phase, \textit{LF exploration}, \mbox{\tool} systematically generates candidate LFs from multiple perspectives guided by task description and data characteristics to encourage diversity. In the second phase, \textit{LF exploitation}, candidate LFs are evaluated and selectively retained based on estimated performance indicators, such as accuracy and coverage, enabling unreliable or redundant heuristics to be filtered out.
The selected LFs are then used to weakly annotate unlabeled data, and the resulting labels are aggregated through probabilistic label models to produce high-quality pseudo-labels for downstream model training.

We evaluated \tool across eleven widely used text classification benchmarks~\cite{alberto2015tubespam, almeida2011contributions, ren2020denoising, malo2014good, krallinger2017overview, huang2024alchemist, guan2025datasculpt}, comparing it against state-of-the-art automated LF generation methods~\cite{huang2024alchemist, guan2025datasculpt, varma2018snuba}. 
\tool consistently produced higher-quality weak labels with near-complete data coverage, leading to substantial gains in downstream model performance.  
On average, \tool achieved a coverage of 98.9\%, notably higher than the other LF generation approaches (ranging from 78.6\% to 95.1\%). 
Moreover, the datasets labeled by \tool exhibit markedly better quality, improving weighted F1 scores by 9--87\% relative to prior LF generation methods. 
These higher-quality labels translated into significant end-to-end (E2E) improvements, with downstream models trained on \tool-labeled data achieving 3--46\% relative gains in weighted F1.
Especially, \tool enhanced label quality by up to 133\% over the baselines on the Yelp Reviews dataset and doubled the downstream performance compared to \alchemist on the ChemProt relation classification task.

In brief, this paper makes the following contributions:
\begin{enumerate}

    \item We propose \mbox{\tool}, an automated framework for programmatic labeling that directly addresses the diversity-reliability gap in existing approaches by systematically \textit{exploring} LFs across surface, structural, and semantic levels, and subsequently \textit{exploiting} them through a principled selection process to ensure a diverse and reliable LF set.

    \item We introduce a principled LF selection and calibration mechanism that filters noisy or redundant heuristics while balancing coverage-precision trade-offs to improve the quality of weak labels ahead of aggregation.
    
    \item We conduct extensive experiments demonstrating that \mbox{\tool} consistently outperforms state-of-the-art automated LF generation approaches in terms of both labeling quality and downstream task performance.

\end{enumerate}
The detailed experimental results and source code for reproducing experiments can be found on our website~\cite{website}.

The remainder of this paper is organized as follows. Section~2 introduces the background on programmatic labeling and formally defines the automated label function generation problem, establishing the notation used throughout the paper. Section~3 presents \tool's framework, including label function exploration, exploitation, and aggregation. Section~4 describes the experimental setup, evaluation metrics, and baseline methods. Section~5 reports and analyzes the experimental results, including ablation studies and practical implications. 
Section~6 reviews relevant work in this field and discusses the limitations of existing approaches.
Finally, Section~7 concludes the paper by summarizing our findings, limitations, and future research directions.
\section{Background and Problem Formulation}
\label{sec:background}

This section introduces the background and the formal problem formulation of automated label function generation. The notation and definitions established in this section provide a unified framework that is used consistently throughout the remainder of the paper.

\subsection{Programmatic Labeling with Weak Supervision}
Supervised machine learning relies on large, high-quality labeled datasets, yet manual annotation is costly and time-consuming.
\textit{Programmatic labeling}, also known as \textit{weak supervision}, alleviates this bottleneck by using \textit{labeling functions}, such as programmatic rules, heuristics, or models, that assign \textit{noisy labels} at scale. A \textit{label model} then aggregates these noisy signals to produce probabilistic training labels for downstream classifiers.

Formally, let $\mathcal{X}$ denote the input space (e.g., sentences in text classification or images in vision tasks), 
and let $\mathcal{Y} = \{1, \dots, C\}$ be the label space for a $C$-class classification problem.   
We are given two datasets:
\begin{itemize}
    \item A large unlabeled dataset 
    $D = \{x_i\}_{i=1}^{N}$, where each $x_i \in \mathcal{X}$ is an input instance.
    
    \item A much smaller labeled seed dataset $D_l = \{(x_i, y_i)\}_{i=1}^{N_l}$, 
    where each $y_i \in \mathcal{Y}$ is a ground-truth label. 
\end{itemize}
Typically, $N \gg N_l$, reflecting the practical situation where labeled data are scarce while unlabeled data are abundant.

A \textit{Labeling Function (LF)} is a programmatic rule or heuristic defined as a mapping $\lambda: \mathcal{X} \to \mathcal{Y} \cup \{\bot\}$, where $\bot$ denotes an \emph{abstention}, meaning the LF does not assign a label to the input. For example, in a sentiment classification task, an LF might assign the label 
\texttt{positive} if the word ``\textit{excellent}'' appears in the text, 
\texttt{negative} if the word ``\textit{terrible}'' appears, and $\bot$ otherwise.

Given a collection of $m$ LFs, denoted $\Lambda = \{\lambda_1, \dots, \lambda_m\}$, the outputs of these LFs on the unlabeled dataset $D$ can be represented as a \textit{label matrix}:
\[
L \in (\mathcal{Y} \cup \{\bot\})^{N \times m}, 
\quad L_{i,j} = \lambda_j(x_i)
\]
where $L_{i,j}$ is the label assigned by LF $\lambda_j$ to instance $x_i$, or $\bot$ if the LF abstains.  
Each row $L_{i,\cdot}$, therefore, contains all LF outputs for a single input $x_i$.

A \textit{label model}, denoted $g_\theta$, is then applied to aggregate the noisy and potentially conflicting LF outputs into a probabilistic label distribution, 
$\hat{p}_i = g_\theta(L_{i,\cdot}) \in \Delta(\mathcal{Y})$,
where $\Delta(\mathcal{Y})$ is the probability simplex over $\mathcal{Y}$. 
For example, for a binary classification task $\mathcal{Y}=\{0,1\}$, 
$\hat{p}_i = (0.8, 0.2)$ indicates that the model estimates $x_i$ has an $80\%$ chance of belonging to class $0$ and $20\%$ to class $1$.

These probabilistic labels $\{\hat{p}_i\}$ are then used to train a \textit{discriminative classifier} $h_\phi: \mathcal{X} \to \Delta(\mathcal{Y})$, which learns to map inputs directly to class probabilities. 
The quality of the final classifier is evaluated on a held-out labeled test set $\mathcal{T}$ using a task-specific metric, denoted by $M(h_\phi; \mathcal{T})$ (e.g., accuracy or F1 score).

\subsection{Automated Label Function Generation}

While programmatic labeling reduces manual annotation, constructing LFs still demands expert effort.

\textbf{Input.} Let $\mathcal{X}$ and $\mathcal{Y}$ denote the input and label spaces, respectively. In addition to the unlabeled dataset $D = \{x_i\}_{i=1}^{N}$ and the small labeled seed set $D_l = \{(x_i, y_i)\}_{i=1}^{N_l}$ with $N_l \ll N$, we are also given a textual task description $t$ specifying the requirements of the task (e.g., sentiment classification).  

\textbf{Objective.} The LF generation task is automatically constructing a set of LFs $\Lambda = \{\lambda_1, \dots, \lambda_m\}$, where each $\lambda_j: \mathcal{X} \to \mathcal{Y} \cup \{\bot\}$ produces a weak label or abstains. These LFs are applied to $D$ to produce the label matrix $L$, which is aggregated by a label model to infer probabilistic labels, then used to train a discriminative classifier. Formally, the objective is to find the optimal set of LFs, $\Lambda^\star = \arg\max_{\Lambda} \; M(h_\phi; \mathcal{T})$, where $h_\phi$ is the end classifier trained on probabilistic labels, $\mathcal{T}$ is a held-out test set, and $M$ is the task-specific evaluation metric.






\section{\tool: Structured Exploration and Exploitation of Label Functions}
\label{sec:approach}

\begin{figure*}
    \centering
    \includegraphics[width=\linewidth]{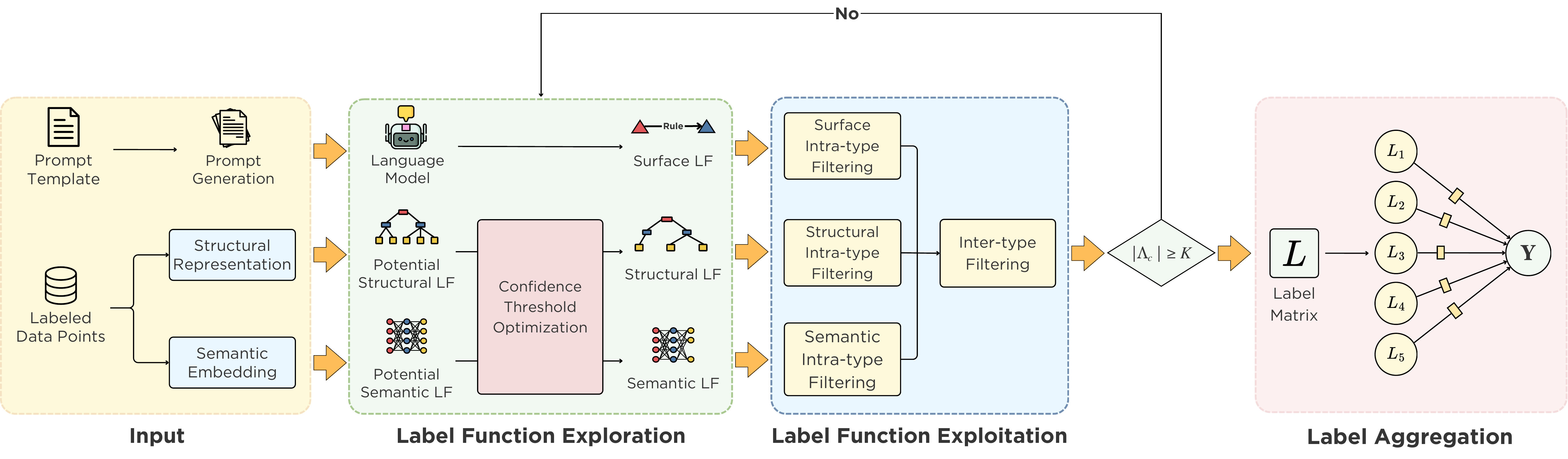}
    \caption{\tool: Approach Overview}
    \label{fig:AutoLF}
\end{figure*}

Figure~\ref{fig:AutoLF} shows an overview of our approach. 
Our main idea in \tool is that the set of LFs should form a \textit{diverse} yet \textit{reliable} ensemble, where diversity ensures broad coverage of \textit{surface}, \textit{structural}, and \textit{semantic} signals, and reliability is enforced both during LF generation and through principled selection to suppress noise.
Particularly, the automated LF generation process in \tool consists of two phases: \textit{label function exploration} and \textit{label function exploitation}.
In the exploration phase, \tool systematically explores candidate LFs that are not only diverse but also guided by LFs' precision to ensure that the generated pool already contains potentially reliable heuristics (Section~\ref{sec:exploration}). 
In the exploitation phase, \tool further refines this pool by selecting high-quality LFs and filtering the underperforming ones based on their estimated accuracy and coverage (Section~\ref{sec:exploitation}). 
The selected LFs are then used to produce weak labels, which are aggregated using weak supervision techniques (Section~\ref{sec:aggregation}) and used to train downstream models.

\subsection{Label Function Exploration}
\label{sec:exploration}

A central challenge in programmatic labeling is to design a diverse and complementary set of LFs that can cover different aspects of the input space. 
In this work, \tool introduces a principled taxonomy of three LF categories: \textit{surface}, \textit{structural}, and \textit{semantic}.  
This taxonomy reflects increasing levels of abstraction in textual data, ranging from direct surface cues to corpus-level structural patterns and deep semantic representations.  
By systematically exploring LFs across these three complementary families, \tool ensures both diversity of supervision signals and robustness of coverage.  
Each LF category embodies a different inductive bias: surface LFs rely on observable lexical patterns, structural LFs exploit corpus-level distributions and statistical regularities, while semantic LFs capture contextualized meaning through pretrained representations.

\subsubsection{Surface Label Function}
Surface LFs represent the most direct form of weak supervision, relying on observable lexical or syntactic signals/cues in the raw input. 
The core principle behind surface LFs is that certain words, phrases, or simple textual patterns often provide strong but narrow signals of class membership. 
Formally, a surface LF $\lambda_{\text{surf}}: \mathcal{X} \to \mathcal{Y} \cup \{\bot\}$ can be defined by a finite set of lexical patterns $\mathcal{K} \subset \mathcal{V}$, where $\mathcal{V}$ denotes the vocabulary. 
The LF outputs a label $y \in \mathcal{Y}$ if an element of $\mathcal{K}$ is detected in $x \in \mathcal{X}$, and $\bot$ (abstain) otherwise:
\[
\lambda_{\text{surf}}(x) = 
\begin{cases}
    y, & \text{if } \exists t \in \mathcal{K}: t \in x, \\
    \bot, & \text{otherwise}.
\end{cases}
\]



\begin{figure}
    \centering
    \includegraphics[width=1\linewidth]{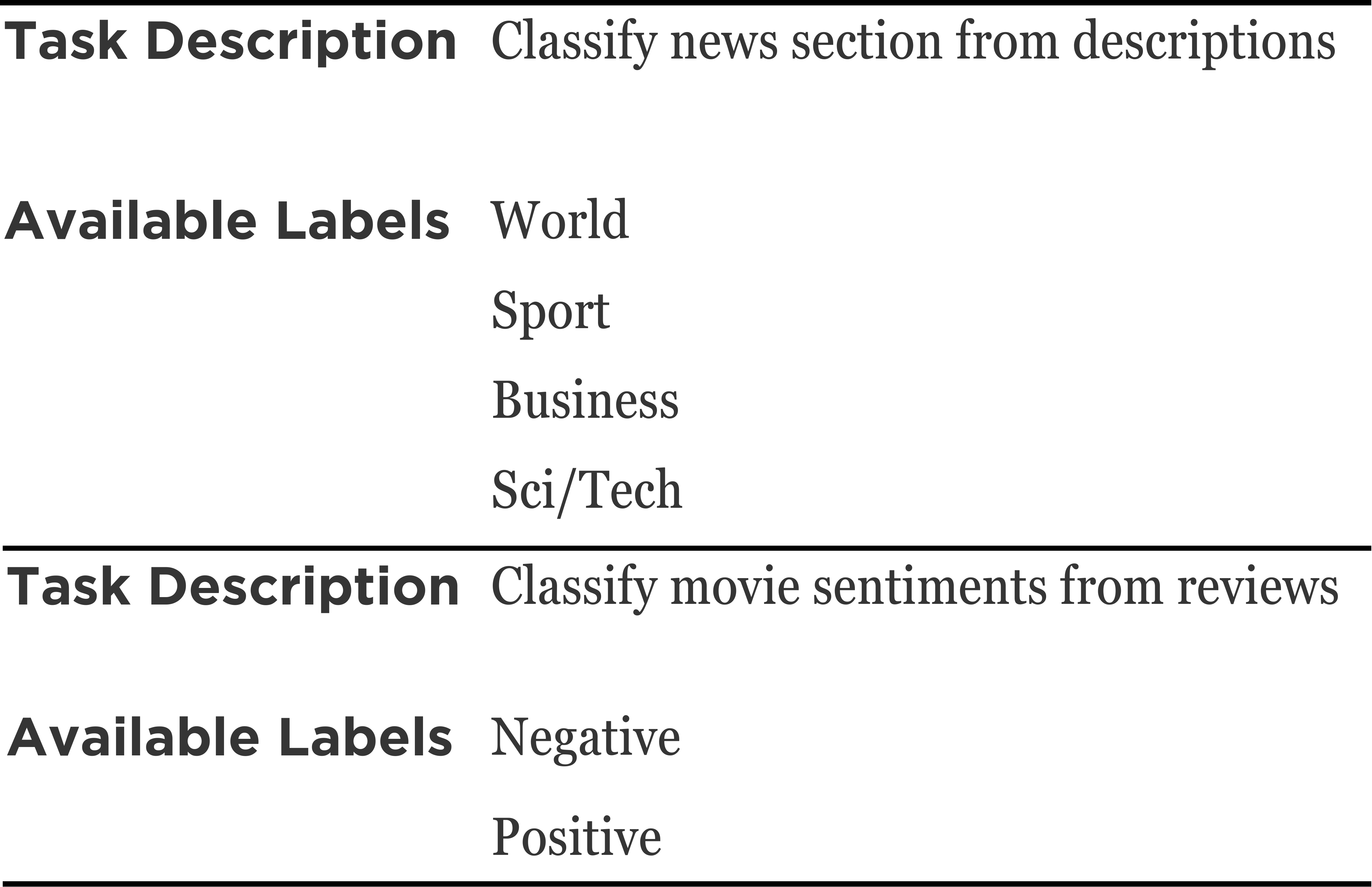}
    \caption{Prompt templates of \tool for surface label function exploration.}
    \label{fig:prompt_template}
\end{figure}

In \tool, we automate the process of generating surface LFs, which are executable labeling rules by leveraging LLMs. We design a prompting framework that instructs an LLM to generate labeling rules representing simple programs given task-specific context, as illustrated in Figure~\ref{fig:prompt_template}.

To support diverse tasks, we design a general and extensible prompt template that contains two main components:
\begin{itemize}
    \item \textbf{Task Description:} a structured summary of the dataset domain and input format, providing the LF generator with context about the nature of the inputs.
    
    \item \textbf{Available Labels:} explicit specification of the label space $\mathcal{Y}$, the order of provided labels is automatically mapped to zero-based numeric labels respectively. 
\end{itemize}

\noindent Surface LFs typically achieve high precision when their cues often suffer from limited coverage and poor domain transferability.
These limitations are effectively mitigated by the complementary structural and semantic LFs, whose generation processes are discussed in the following sections.



\subsubsection{Structural Label Function}
\label{sec:struct_lf}

Structural LFs aim to capture statistical and syntactic regularities that go beyond local lexical cues.
These LFs are designed to (i) exploit frequency and co-occurrence statistics and (ii) be robust to local lexical variation.
In \tool, structural LFs are automatically generated by training lightweight classifiers on limited labeled subsets to exploit frequency and co-occurrence statistics.
Because $N_l$ is small, we employ simple and well-regularized models that can produce calibrated probability estimates and avoid overfitting under low-supervision conditions.

Let $\phi_{\text{tfidf}}:\mathcal{X}\to\mathbb{R}^{d_{\text{tfidf}}}$ denote a TF-IDF feature extractor (e.g., unigram/bigram counts scaled by inverse document frequency). For $k=1,\dots,m$ (number of structural candidates), we construct a classifier 
$f^{k}_{\theta_k} : \mathbb{R}^{d_{\text{tfidf}}} \to \Delta(\mathcal{Y})$
trained on a random subsample $S^{k} \subseteq D_l$ (sampled without replacement) of size $s < N_l$. Training minimizes an empirical loss (e.g., hinge loss for an SVM with probability calibration) to yield parameters $\theta_k$.
The classifier outputs a predictive distribution $p^{k}(y\mid x) = f^{k}_{\theta_k}\big(\phi_{\text{tfidf}}(x)\big)$, $y\in\mathcal{Y}$.

In general, $f^{k}_{\theta_k}$ could directly used as a label function by considering $\arg\max_{y\in\mathcal{Y}} p^{k}(y\mid x)$ as predicted weak labels. However, since $f^{k}_{\theta_k}$ is trained on a limited labeled set ($D_l$), its predictions may suffer from high variance and poor generalization, leading to unreliable weak labels. Consequently, these unreliable predictions can result in incorrect predicted labels.
To mitigate the effect of $f^{k}_{\theta_k}$'s unreliability, we transform it into a structural LF $\lambda^k$ by introducing a confidence threshold $\omega_k \in [0,1]$, ensuring that only sufficiently confident predictions are used as weak labels:
\[
\label{eq:lf-map}
\lambda_{\text{struct}}^{k}(x) =
\begin{cases}
\arg\max_{y\in\mathcal{Y}} p^{k}(y\mid x) & \text{if } \max_{y} p^{k}(y\mid x) > \omega_k,\\
\bot & \text{otherwise.}
\end{cases}
\]
Thus, each candidate provides a probabilistic vote when sufficiently confident and abstains otherwise. The impact of this technique on \tool's performance will be experimentally discussed in Section~\ref{sec:results}.

For each structural candidate, we select $\omega_k$ by optimizing a trade-off between \textit{precision} and \textit{coverage}. 
The design objective of LFs is to produce supervision signals that are both reliable and broadly applicable, rather than aiming for perfect instance-level accuracy. Setting $\omega_k$ too high yields overly cautious LFs with limited coverage, while too low a threshold increases labeling noise. Hence, optimizing $\omega_k$ entails balancing precision and coverage so that each LF provides sufficiently confident pseudo-labels without becoming overly restrictive. This calibration ensures that LFs meaningfully contribute to the label model by improving consensus quality, aligning with the weak supervision principle of aggregating diverse yet trustworthy signals.

To perform this trade-off, we estimate the precision and coverage of $\lambda^k$ as functions of $\omega$ over the labeled subset $D_l$. Specifically, the \textit{precision} of the LF $\lambda^{k}$ parameterized by threshold $\omega$ (denoted $\lambda^{k}_\omega$) on $D_l = \{(x_i, y_i)\}_{i=1}^{N_l}$ is formally defined as: 
\[
\mathrm{prec}_k(\omega) =
\frac{\sum_{(x,y)\in D_l} \mathbf{1}\big[\lambda^{k}_\omega(x)=y\big]}
{\sum_{(x,y)\in D_l} \mathbf{1}\big[\lambda^{k}_\omega(x)\neq\bot\big]+\epsilon},
\]
with a small $\epsilon>0$ to avoid division by zero.

The \textit{coverage} of $\lambda^{k}_\omega$ on the labeled dataset $D_l = \{x_i, y_i\}_{i=1}^{N_l}$ is defined as:
\[
\mathrm{cov}_k(\omega) = \frac{1}{N_l}\sum_{(x,y)\in D_l} \mathbf{1}\big[\lambda^{k}_\omega(x)\neq\bot\big]
\]
When $D_l$ is small, coverage can alternatively be estimated over the unlabeled set $D$, providing a more stable approximation.
We choose $\omega_k^\star$ by maximizing a Weighted Harmonic Mean between precision and coverage:
\[
\omega_k^\star = \arg\max_{\omega \in [0,1]} \mathrm{WHM}\!\left(\mathrm{prec}_k(\omega), \mathrm{cov}_k(\omega)\right),
\]
where the weighted harmonic mean is defined as
\[
\mathrm{WHM}\!\left(\mathrm{prec}_k(\omega), \mathrm{cov}_k(\omega)\right)
= \frac{(1+\beta^2)\,\times\mathrm{prec}_k(\omega)\,\times\mathrm{cov}_k(\omega)}
{\beta^2\,\times\mathrm{prec}_k(\omega) + \mathrm{cov}_k(\omega)}
\]

This penalizes extreme settings (very high precision but negligible coverage, or vice versa) and aligns with the exploration goal of producing useful, reliable candidates. Empirically, we find that coverage is generally easier to achieve than precision, motivating $\beta \ll 1$ to place greater emphasis on precision improvements during threshold selection. 

To increase diversity among the $m$ candidates, one could vary: the subsample $S^{k}$, the TF-IDF $n$-gram ranges, regularization hyper-parameters, and feature selection seeds. Each such variation yields a structurally distinct LF that may capture different corpus regularities.

\subsubsection{Semantic Label Function}

Semantic LFs are designed to capture contextual and distributional meaning that surface and structural cues may miss. They encode sentence- or document-level semantics via dense representations derived from pretrained language models, and produce predictions that generalize across lexical variation. Given the limited labeled seed, we favor \textit{frozen-embedding + lightweight classifier} designs to limit overfitting while preserving semantic generalization.

Let $\phi_{\text{sem}}:\mathcal{X}\to\mathbb{R}^{d_{\text{sem}}}$ denote a contextual embedding extractor (e.g., pooled outputs from a pretrained transformer). For $k=1,\dots,m_{\text{sem}}$ we train a shallow classifier $g^{(k)}_{\psi_k} : \mathbb{R}^{d_{\text{sem}}} \to \Delta(\mathcal{Y})$, on a random subsample $S^{(k)}\subseteq D_l$. Typically, the underlying embedding extractor is \textit{fixed} to reduce labeled-data requirements; only the parameters of the classifier head $\psi_k$ are trained. 
The predicted distribution is $q^{(k)}(y\mid x) = g^{(k)}_{\psi_k}\big(\phi_{\text{sem}}(x)\big)$.

As with structural LFs, each semantic candidate is converted to an LF by thresholding:
\[
\lambda^{(k)}_{\text{sem}}(x) =
\begin{cases}
\arg\max_y q^{(k)}(y\mid x) & \text{if }\max_{y} q^{(k)}(y\mid x) > \omega_k',\\
\bot & \text{otherwise.}
\end{cases}
\]
The threshold $\omega_k'$ is chosen by the same weighted harmonic-mean procedure described in Section~\ref{sec:struct_lf} (coverage on $D$ vs.\ precision on $D_l$), producing $\omega_k'^\star$.

To populate the semantic candidate pool, one could vary: the embedding pooling strategy (e.g., mean pooling, layer aggregation), the Multilayer Perceptron (MLP) width/depth, the subsample $S^{(k)}$, and random initialization seeds. When feasible, we also consider lightweight fine-tuning variants (a small number of gradient steps) as additional candidates, but these are included only if validation indicates they do not overfit the seed.

\textbf{Summary}. Applying the procedures above provides a candidate pool
\[
\Lambda = \big\{ \lambda^{(k)}_{\text{surf}} \big\}_{k} \cup\ \big\{ \lambda^{(k)}_{\text{struct}} \big\}_{k} \cup \big\{ \lambda^{(k)}_{\text{sem}} \big\}_{k},
\]
where each candidate is a mapping $\mathcal{X}\to\mathcal{Y}\cup\{\bot\}$ with an associated confidence threshold chosen to balance coverage and precision. This heterogeneous pool is the input to the exploitation (selection and filtering) phase, which ranks and prunes candidates before label aggregation.

\subsection{Label Function Exploitation}
\label{sec:exploitation}

The goal of the LF exploitation phase is to refine the automatically generated pool of label functions into a subset that is both \textit{diverse} and \textit{reliable}, maximizing downstream label quality.
While the exploration phase emphasizes breadth by systematically generating candidate LFs across surface, structural, and semantic categories, the exploitation phase emphasizes depth by filtering out noisy or redundant signals. 
This step is essential as weak supervision relies on the interplay of overlapping, conflicting, and abstaining LFs. Thus, excessive noise can give misleading signals to the label model's estimates of LF accuracies and correlations.

Formally, let $\Lambda = \{\lambda_1, \dots, \lambda_m\}$ denote the pool of LFs generated in the exploration phase. 
For each $\lambda_j \in \Lambda$, we estimate its accuracy $\widehat{acc}(\lambda_j)$ using the small labeled dataset $D_l$.
These estimated accuracies are then used to filter out underperforming or inconsistent LFs. Specifically, \tool selects a refined subset $\Lambda' \subseteq \Lambda$ by (i) removing LFs whose $\widehat{acc}(\lambda_j)$ falls below an adaptive threshold (\textit{intra-type filtering}) and (ii) enforcing category-level diversity to maintain balanced representation among surface, structural, and semantic LFs (\textit{inter-type filtering}).
The resulting $\Lambda'$ thus provides a compact yet expressive ensemble that achieves high coverage while minimizing noise, ensuring reliable weak label aggregation in subsequent steps.


\textbf{Intra-type filtering}. Within each LF category, we estimate the accuracy of individual label functions. Any LF whose estimated accuracy falls below a threshold $\theta_1$ is discarded. The threshold is defined as
\[
    \theta^c_{\text{intra}} = \alpha \cdot \max_{\lambda_j \in \Lambda_c} \widehat{acc}(\lambda_j)
\]
where $\Lambda_c$ is the LF set from category $c$, either surface, structural, or semantic, and $\alpha \in (0,1]$ is a predefined acceptance multiplier. This formulation allows the threshold to adapt to the performance distribution within each category, retaining high-quality LFs while discarding underperforming ones. 

\textbf{Inter-type filtering}. After intra-type filtering, the remaining LFs from all categories are jointly evaluated to remove globally weak candidates. To maintain a balance between quality and diversity, we apply a relaxed global threshold defined as:
\[
\theta_{\text{inter}} = \tfrac{1}{2} \max \{\theta_{\text{intra}}^{\text{surface}}, \theta_{\text{intra}}^{\text{structural}}, \theta_{\text{intra}}^{\text{semantic}}\}
\]
This inter-type filtering stage serves two purposes. First, it prevents low-quality LFs that survive within-category filtering from degrading the overall labeling performance. Second, by using a moderated threshold (half of the highest intra-type threshold), it avoids over-penalizing LF types with inherently lower estimated accuracies.
This design stage acknowledges that different LF types capture complementary and heterogeneous aspects of data and should not be directly compared with identical strictness. The softer criterion eliminates only extreme underperformers, thus preserving diversity across categories.

\textbf{Loop-Back for Iterative LF Discovery}.
Through this two-stage filtering, \tool balances \textit{accuracy} (removing low-quality LFs) and \textit{diversity} (retaining heterogeneous perspectives). 
The resulting set $\bar{\Lambda}$ forms a high-quality supervision source for the aggregation phase, where the label model leverages overlaps, conflicts, and abstentions to generate probabilistic labels for the unlabeled dataset $D$.

To further enhance coverage and category balance, \tool incorporates a loop-back mechanism that re-invokes the exploration phase after exploitation. 
Specifically, if the number of high-quality LFs in any category $c \in \{\text{surface}, \text{structural}, \text{semantic}\}$ falls below a predefined target $K_c$, the framework triggers a new round of LF exploration focused on that category.
In each iteration, the generation process is guided by diagnostic feedback, such as coverage gaps, disagreement patterns, or low consensus regions, identified during exploitation. 
To avoid redundancy, previously accepted LFs are embedded and compared against newly proposed ones using category-specific similarity measures (e.g., lexical overlap for surface, feature correlation for structural, and embedding similarity for semantic), ensuring that subsequent rounds introduce genuinely novel heuristics.
This iterative refinement enables \tool to progressively discover new, complementary labeling heuristics while maintaining a balanced and diverse LF set across categories. 
The process continues until all categories meet their target LF counts ($|\bar{\Lambda}_c| \ge K$ for all $c$), ensuring both sufficient coverage and robustness before proceeding to label aggregation.

Algorithm~\ref{alg:algorithm} comprehensively summarizes our process of automatic label function generation in \tool.
The method initializes three category-specific LF pools and iteratively generates candidate LFs for surface, structural, and semantic types until each category reaches the predefined maximum $K$ (line~4).
Structural and semantic LFs are further calibrated using the labeled set $D_l$ by estimating precision, coverage, and optimizing confidence thresholds (lines~7--12) in the Algorithm~\ref{alg:algorithm}.
Newly generated LFs are then refined through intra-type filtering (lines~14--16) to remove weak or redundant candidates, followed by an inter-type filtering step to ensure cross-category quality and diversity (lines~18 and 19).
The final candidate LF pool $\Lambda$ is obtained by aggregating the filtered LFs from all categories.

\subsection{Weak-label Aggregation}
\label{sec:aggregation}

Individual LFs provide useful but noisy supervision signals, often with limited coverage. However, surface, structural, and semantic LFs contribute complementary strengths, making their combination more effective than any single source. 
To integrate these heterogeneous signals, we adopt weak supervision techniques~\cite{ratner2017snorkel}, which learn a generative label model to estimate LF accuracies and correlations. The label model then aggregates multiple noisy labels into probabilistic pseudolabels of higher quality.
Our framework is compatible with various weak supervision methods, but for simplicity and reproducibility, we employ the widely adopted \snorkel framework~\cite{ratner2017snorkel}. This choice ensures both scalability and comparability with prior work.

\begin{algorithm}[ht]
\small
\SetAlgoLined
\DontPrintSemicolon
\SetInd{0.5em}{0.8em}

\SetKwProg{Proc}{Procedure}{}{}
\SetKwFunction{ALG}{LabelFunctionGeneration}

\caption{Automatic Label Function Generation of \tool}
\label{alg:algorithm}

\Proc{\ALG{$D_l, D$}}{
\KwIn{Labeled data $D_l$, Unlabeled data $D$}
\KwOut{Candidate pool $\Lambda$}

\Begin{
$\{\lambda_{\text{surf}}\}, \{\lambda_{\text{struct}}\}, \{\lambda_{\text{sem}}\} \gets \emptyset, \emptyset, \emptyset$\;

\While{$\min(|\{\lambda_{\text{surf}}\}|, |\{\lambda_{\text{struct}}\}|, |\{\lambda_{\text{sem}}\}|) < K$}{
    \For{$c \in \{\text{surf, struct, sem}\}$}{
        
        $\{\lambda\}' \gets \textit{generateLFs}(c, K - |\{\lambda_{c}\}|)$\;

        \If{$c \in \{\text{struct, sem}\}$}{
            \For{$\lambda \in \{\lambda\}'$}{
                $\lambda.precision \gets \textit{estimatePrec}(\lambda, D_l)$\;
                $\lambda.coverage \gets \textit{estimateCov}(\lambda, D_l)$\;
                $\lambda.threshold \gets \textit{optimizeThr}(\lambda, D_l)$\;
            }
        }
        $\{\lambda_{c}\}.add(\Lambda')$\;
        $\theta^c_{intra} \gets \textit{intraThreshold}(\{\lambda_{c}\})$\;
        $\{\lambda_{c}\}.\textit{remove}(\theta^c_{intra})$\;
    }

    $\theta_{inter} \gets \textit{interThreshold}(\{\lambda_{\text{surf}}\}, \{\lambda_{\text{struct}}\}, \{\lambda_{\text{sem}}\})$\;

    $\{\lambda_{\text{surf, struct, sem}}\}.\textit{remove}(\theta_{inter})$\;


}
$\Lambda \gets \{\lambda_{\text{surf}}\} \cup \{\lambda_{\text{struct}}\} \cup \{\lambda_{\text{sem}}\}$\;

\KwRet $\Lambda$;
}
}
\end{algorithm}
\section{Evaluation Methodology}
\label{sec:eval_method}

To evaluate the effectiveness of \tool, we aim to address the following research questions:

\begin{itemize}
    \item \textbf{RQ1: Performance Comparison.} How effective is \tool in producing high-quality LFs and training downstream models? How does it compare with state-of-the-art weak supervision methods~\cite{guan2025datasculpt, huang2024alchemist, varma2018snuba}, and semi-supervised learning~\cite{wang2006label}?

    \item \textbf{RQ2: Contribution of LF Categories.} What are the respective roles of \textit{surface}, \textit{structural}, and \textit{semantic} LFs, and how does their combination contribute to overall performance?
    
    \item \textbf{RQ3: Intrinsic Analysis.} How do the key components (i.e., \textit{exploration}, \textit{exploitation}, and \textit{aggregation}) and \tool's parameters affect the final outcome?
    
    \item \textbf{RQ4: Sensitivity Analysis.} How does the amount of labeled data points influence the overall performance of our approach?

    \item \textbf{RQ5: Efficiency Analysis.} How efficient is \tool in terms of execution time and computational cost?

\end{itemize}

\subsection{Datasets}
\begin{table}\centering
\caption{Dataset overview}
\label{tab:datasets}
\begin{tabular}{lrrrrr}\toprule 
Dataset     & \#classes       & $|D|$     & $|D_l|$   & $|D_{test}|$ \\\midrule
YouTube     & 2         & 1,586     & 18        & 250 \\
SMS         & 2         & 4,135     & 77        & 517 \\
IMDb        & 2         & 20,000    & 375       & 2,500 \\
Yelp        & 2         & 30,400    & 570       & 3,800 \\
Clickbait   & 2         & 25,600    & 480       & 3,200 \\
AGNews      & 4         & 96,000    & 1800      & 12,000 \\
Finance     & 3         & 4,673     & 87        & 585 \\
Chemprot    & 10        & 12,861    & 241       & 1,607 \\
Massive     & 18        & 12,021    & 337       & 2,250 \\
PubMed      & 14        & 40,000    & 750       & 5,000 \\
PaperAbs    & 6         & 16,777    & 503       & 839 \\
\bottomrule
\end{tabular}
\end{table}

{Table\mbox{~\ref{tab:datasets}} summarizes the statistics of the datasets used in our evaluation.} Particularly, we evaluated \tool on eleven datasets covering different classification scenarios, which are widely used in the existing studies in data labeling~\cite{huang2024alchemist,guan2025datasculpt,wang2006label,weaker-than-you-think}.
{These datasets span a wide range of domains,} including social media and spam detection (YouTube, SMS, Clickbait), user sentiment reviews (IMDb, Yelp, Finance), topic categorization (AGNews, PaperAbs), as well as specialized domains such as biomedical literature (Chemprot, PubMed). 
This diversity ensured a comprehensive evaluation of our method across binary (YouTube, SMS, IMDb, and Yelp), multiclass (AGNews, Clickbait, Finance, Chemprot, and Massive), and multilabel (PubMed and PaperAbs) classification tasks. 
For each dataset, we randomly sampled about 1.5\% of the original dataset to form a labeled seed set $D_l$, with the remaining data treated as the unlabeled set $D$. 
For datasets where an official test split $D_{test}$ is not provided, we randomly sampled 10\% of the original dataset as a held-out gold test set $D_{test}$.
This random sampling procedure did not explicitly enforce class-stratified splits; therefore, the class distribution of $D_{test}$ approximately reflects that of the original dataset but may exhibit minor variation, particularly for underrepresented classes.

\subsection{Evaluation Procedure}
\textbf{RQ1. Performance Comparison:}
%
We evaluated \tool against several state-of-the-art and representative baselines:

\begin{itemize}
    \item \textbf{\alchemist}~\cite{huang2024alchemist}: a framework that leverages LLMs to generate executable labeling programs rather than directly annotating data. This approach enables reusability and significant cost savings while maintaining competitive label quality.
    
    \item \textbf{\datasculpt-base}~\cite{guan2025datasculpt}: a method that employs few-shot prompting of LLMs within a programmatic weak supervision pipeline. It produces diverse LFs through iterative expansion, achieving high labeling accuracy with reduced annotation cost.

    \item \textbf{\snuba}~\cite{varma2018snuba}: a model-based heuristic synthesis approach that generates simple LFs from a small labeled set and user-defined primitives. It iteratively expands coverage while maintaining label accuracy by abstaining when uncertain and stopping once additional heuristics no longer improve quality.

    \item \textbf{Few-shot Learning}: a \textit{shallow} feedforward neural network trained solely on the small labeled subset $D_l$, representing a standard supervised learning setting under limited supervision.

    \item \textbf{Label propagation}: a classical semi-supervised baseline using label propagation~\cite{wang2006label} inferring labels for the unlabeled data $D$ by propagating similarities from the seed set $D_l$.
\end{itemize}
All methods were assessed along two complementary dimensions:  
(1) \textit{label quality}, quantified by dataset coverage and labeling accuracy of the generated weak labels; and  
(2) \textit{downstream performance}, measured by training discriminative models on the weakly labeled data and assessing task-specific metrics $M(h_\phi;\mathcal{T})$ on held-out test sets.

\textbf{RQ2. Contribution of LF Categories:}
We analyzed the contribution of individual LF categories (Surface, Structural, and Semantic) in isolation. 
For each setting, \tool was configured to generate and use \textit{only one} LF category during the exploration phase, while all other LF types were disabled.
To ensure that performance differences reflect the intrinsic contribution of each LF category rather than selection effects, the inter-type filtering step was disabled and no cross-category pruning was applied; in other words, LF exploitation was restricted to intra-category processing only.
Experiments were performed on three representative datasets: IMDb (casual comments), Finance (domain-specific text), and Massive (multi-intent classification). This controlled setup enabled us to examine how exploiting LFs at varying levels of linguistic and semantic abstraction improved label quality and downstream performance.


\textbf{RQ3. Intrinsic Analysis:}
We investigated the impact of key components on overall performance. 
For exploration, we built and compared the variants of \tool varied by LLM used to generate surface LFs, classifier used to generate structural LFs, pretrained model to generate semantic LFs, abstain allowance in LFs, and other configuration options/parameters in the phase.
For exploitation, we varied the value of acceptance multiplier $\alpha$ and the maximum number of LFs for each category $K_c$ in \tool to understand their impact on the overall performance.
For aggregation, we compared simple strategies (majority voting, weighted voting) and advanced methods such as FlyingSquid~\cite{fu2020fast}, Dawid-Skene~\cite{dawid1979maximum}, and Snorkel~\cite{ratner2017snorkel}. 


\textbf{RQ4. Sensitivity Analysis:}
We evaluated the effect of labeled data availability on our approach's ability to generate reliable labels for unlabeled instances. Specifically, we varied the proportion of labeled data from 10\% to 100\% of the development set, which corresponds to only about 1\% to 10\% of the entire dataset. This setup highlighted the effectiveness of our method even under limited supervision.



\subsection{Evaluation Metrics}  

To comprehensively evaluate the effectiveness of \tool, we adopted three complementary metrics: \textit{Coverage}, \textit{Labeling Quality}, and \textit{End-to-End (E2E) Performance}.
These metrics jointly assess the framework's ability to (i) annotate unlabeled data effectively, (ii) produce high-quality labels, and (iii) improve downstream model performance.

\textit{Coverage} quantifies the proportion of instances that receive at least one weak label from the set of LFs. High coverage indicates that the framework is able to annotate a substantial portion of the unlabeled dataset. Formally, let $N$ denote the total number of instances, and let $\mathbf{1}_{i}$ be an indicator variable such that $\mathbf{1}_{i}=1$ if instance $x_i$ receives at least one weak label and $\mathbf{1}_{i}=0$ otherwise. Coverage is then defined as:  

$$\text{Coverage} = \frac{1}{N} \sum_{i=1}^{N} \mathbf{1}_{i}$$

\textit{Labeling Quality.} 
Labeling quality evaluates the correctness of labels produced after LF aggregation. It is quantified using the Coverage-Weighted F1-score, which accounts for class imbalance by weighting each class-specific F1 by its true class proportion. Let $\mathcal{Y}$ be the set of classes, $N_c$ the number of true instances of class $c$, and $N = \sum_{c \in \mathcal{Y}} N_c$.  
For each class $c$, the class-specific F1 is:  

$$\textit{F1}_c = 2 \times \frac{Precision_c \times Recall_c}{Precision_c + Recall_c}$$
\noindent The weighted F1-score is then defined as:  

$$\textit{F1}_{\textit{weighted}} = \sum_{c \in \mathcal{Y}} \frac{N_c}{N} \times \textit{F1}_c$$
\noindent To reflect both correctness and coverage, we compute:

$$
\text{Label. Quality} = \text{Coverage} \times \textit{F1}_{\textit{weighted}}
$$
\noindent This formulation ensures that the overall labeling quality decreases proportionally when the framework labels fewer instances, even if per-class precision and recall are high. 

\textit{End-to-End (E2E) Performance.}
E2E performance assesses how well a downstream classifier trained solely on the weakly labeled dataset performs on the held-out gold test set.
Like labeling quality, it is measured using the standard Weighted F1-score, providing a consistent basis for comparison between labeling and downstream learning stages.

\subsection{Experiment Setup}
For all automated data annotation techniques, we employed an identical downstream model architecture, which is an MLP with one hidden layer, containing 100 units, utilizing the ReLU function. All experiments were conducted 5 times with different seeds, and their average scores are reported.
In the default setting, \tool used GPT-4.1 to generate surface LFs, SVM in generating structural LFs, BERT-base in generating semantic LFs, $\beta = 0.1$, and $\alpha = 0.9$. We sat $20$ as the maximum number of LFs generated by \tool for each category.

In \tool, we leveraged popular libraries: LangChain (version 0.3.26) to interact with LLMs for generating lexical LFs, and scikit-learn (version 1.2.2) to construct the other two types of LFs (statistical and semantic). For embeddings, we adopted BERT~\cite{devlin2019bert} models obtained from HuggingFace. All experiments were conducted on a Linux 5.15.154 server equipped with two NVIDIA T4 GPUs.
\section{Experiments}
\label{sec:results}

\subsection{Performance Comparison}

\begin{table*}
\centering
\footnotesize
\caption{Labeling performance comparison on binary- (YouTube, SMS, IMDb, Yelp, and Clickbait) and multi-class (the others) datasets.}
\label{tab:single_label}
\begin{tabular}{l|l|ccccc|ccccc}\toprule
& &\textit{YouTube} &\textit{SMS} &\textit{IMDb} &\textit{Yelp} &\textit{Clickbait} &\textit{Agnews} &\textit{Finance} &\textit{Chemprot} &\textit{Massive} \\\midrule
\multirow{4}{*}{\textbf{\#LFs}}
&\snuba &25 &3 &25 &25 &25 &-- &-- &-- &-- \\
&\alchemist &15 &15 &15 &15 &15 &15 &15 &15 &15 \\
&\datasculpt &84 &167 &90 &78 &171 &216 &185 &181 &131 \\
&\tool &60 &60 &60 &60 &60 &60 &58 &46 &60 \\\midrule
\multirow{4}{*}{\textbf{Coverage}}
&\snuba &1.000 &1.000 &0.605 &0.650 &0.185 &-- &-- &-- &-- \\
&\alchemist &1.000 &0.999 &0.982 &0.928 &0.635 &0.996 &1.000 &1.000 &0.935 \\
&\datasculpt &0.796 &0.598 &0.960 &0.941 &0.487 &0.721 &0.623 &0.887 &0.698 \\
&\tool &1.000 &1.000 &1.000 &1.000 &1.000 &1.000 &1.000 &0.930 &0.945 \\\midrule
\multirow{4}{*}{\textbf{Weighted F1}}
&\snuba &0.630 &0.461 &0.708 &0.779 &0.895 &-- &-- &-- &-- \\
&\alchemist &0.724 &0.944 &0.781 &0.829 &0.834 &0.577 &0.570 &0.338 &0.632 \\
&\datasculpt &0.623 &0.859 &0.740 &0.394 &0.827 &0.648 &0.442 &0.304 &0.411 \\
&\tool &0.848 &0.962 &0.789 &0.865 &0.959 &0.871 &0.623 &0.509 &0.720 \\\midrule
\multirow{6}{*}{\textbf{Label. Quality}}
&\fewshot &0.711 &\ul{0.950} &0.715 &\ul{0.783} &\ul{0.931} &\ul{0.808} &0.478 &\ul{0.369} &0.532 \\
&\labelprop &0.619 &0.920 &0.568 &0.633 &0.856 &0.591 &\ul{0.590} &0.281 &0.314 \\
&\snuba &0.630 &0.461 &0.428 &0.506 &0.166 &-- &-- &-- &-- \\
&\alchemist &\ul{0.724} &0.943 &\ul{0.767} &0.769 &0.530 &0.575 &0.570 &0.338 &\ul{0.591} \\
&\datasculpt &0.496 &0.514 &0.710 &0.371 &0.403 &0.467 &0.275 &0.270 &0.287 \\
&\tool &\textbf{0.848} &\textbf{0.962} &\textbf{0.789} &\textbf{0.865} &\textbf{0.959} &\textbf{0.871} &\textbf{0.623} &\textbf{0.455} &\textbf{0.680} \\\midrule
\multirow{6}{*}{\textbf{E2E Perf.}}
&\fewshot &0.713 &\ul{0.962} &0.724 &0.817 &\ul{0.937} &\ul{0.805} &0.504 &\ul{0.355} &\ul{0.533} \\
&\labelprop &0.695 &0.917 &0.559 &0.643 &0.849 &0.609 &\ul{0.583} &0.282 &0.307 \\
&\snuba &0.778 &0.461 &0.697 &0.773 &0.932 &-- &-- &-- &-- \\
&\alchemist &\ul{0.823} &0.955 &0.729 &\ul{0.841} &0.923 &0.693 &0.544 &0.178 &0.483 \\
&\datasculpt &0.658 &0.865 &\ul{0.738} &0.366 &0.931 &0.773 &0.491 &0.230 &0.325 \\
&\tool &\textbf{0.870} &\textbf{0.976} &\textbf{0.738} &\textbf{0.853} &\textbf{0.958} &\textbf{0.827} &\textbf{0.617} &\textbf{0.365} &\textbf{0.575} \\
\bottomrule
\end{tabular}
\end{table*}

\begin{table}
\centering
\caption{Labeling performance comparison on multi-label datasets.}
\label{tab:multi_label}
\footnotesize
\begin{tabular}{l|l|cccc}\toprule
& &\textit{Pubmed} &\textit{PaperAbs} \\\midrule
\multirow{3}{*}{\textbf{\#LFs}} 
&\alchemist &210 &90 \\
&\datasculpt &6241 &2679 \\
&\tool &772 &332 \\\midrule
\multirow{3}{*}{\textbf{Coverage}} 
&\alchemist &1.000 &0.986 \\
&\datasculpt &0.963 &0.972 \\
&\tool &1.000 &1.000 \\\midrule
\multirow{3}{*}{\textbf{Weighted F1}} 
&\alchemist &0.762 &0.819 \\
&\datasculpt &0.629 &0.728 \\
&\tool &0.820 &0.886
\\\midrule
\multirow{5}{*}{\textbf{Label. Quality}} 
&\fewshot &\ul{0.808} &\ul{0.850} \\
&\labelprop &0.720 &0.788 \\
&\alchemist &0.762 &0.807 \\
&\datasculpt &0.606 &0.707 \\
&\tool &\textbf{0.820} &\textbf{0.886} \\\midrule
\multirow{5}{*}{\textbf{E2E Perf.}} 
&\fewshot &\ul{0.806} &\ul{0.845} \\
&\labelprop &0.717 &0.781 \\
&\alchemist &0.766 &0.810 \\
&\datasculpt &0.685 &0.753 \\
&\tool &\textbf{0.806} &\textbf{0.869} \\
\bottomrule
\end{tabular}
\end{table}

%

{Tables\mbox{~\ref{tab:single_label}} and\mbox{~\ref{tab:multi_label}} show the labeling performance of the state-of-the-art programmatic labeling approaches} (\snuba, \alchemist, \datasculpt), as well as classical weakly supervised (\fewshot), and semi-supervised baselines (\labelprop) in terms of labeling quality and downstream performance. 
Overall, \tool consistently outperformed all baselines across binary, multi-class, and multi-label datasets. 
In particular, \tool achieved the best weighted F1-scores while maintaining near-perfect coverage across all datasets. This indicates high labeling quality derived from both the quantity and quality of labeled data points. As a result, downstream task performance was significantly improved. Note that the weighted F1-scores of \fewshot and \labelprop are omitted from tables~\ref{tab:single_label} and~\ref{tab:multi_label} as they do not operate under the weak supervision setting and therefore exhibit comparable weighted F1-scores and labeling quality.

Compared to \alchemist, \tool matched its high coverage while achieving substantially higher labeling accuracy on nearly all benchmarks. For example, \tool improved labeling accuracy by around 10--15\% relative gains on datasets such as IMDb, Yelp, and AgNews. E2E performance of the downstream models followed the same trend, showing clear and consistent advantages. While \alchemist relies on a relatively small, fixed LF pool, \tool's structured exploration and principled filtering allow it to maintain both diversity and reliability, translating into better downstream generalization.
The gap widened further when compared with \datasculpt. Despite generating an order of magnitude more LFs, \datasculpt's label accuracy and E2E performance lagged behind, in some cases by over 50\% reduction in accuracy (e.g., Finance, Massive) compared to \tool's. This indicates that scaling the LF pool without systematic selection introduces excessive noise, undermining label quality. By contrast, \tool produced fewer but higher-quality LFs, showing that LF precision outweighs sheer quantity.
%
%
While \snuba achieved competitive coverage on relatively simple datasets such as \textit{YouTube} and \textit{SMS}, its coverage became worse on more challenging datasets, especially in \textit{Clickbait}. More critically, the labeling quality of \snuba remained limited, entailing modest E2E performance in all datasets. This can be attributed to the simplicity of the generated heuristic models and their inability to capture heterogeneous levels of textual abstraction.

\fewshot, trained directly on $D_l$, performed strongly in some domains (e.g., SMS, Yelp, PubMed). However, \tool surpassed this method in most cases, especially in challenging domains like YouTube, Finance, and PaperAbs. On average, \tool achieved 5--10\% higher labeling performance while maintaining comparable or better E2E performance, despite relying solely on weak supervision. 
%
{Similarly, \mbox{\labelprop} was consistently outperformed by \mbox{\tool} in both label quality and E2E performance.} The improvements were substantial on more complex datasets (e.g., ChemProt), reflecting that \tool's LF-based strategy is better at capturing patterns than simple propagation.


We further analyzed cases where \tool incorrectly assigned labels and found that the main cause was the limited size of $D_l$, which constrained the generation of high-quality LFs. When $D_l$ was expanded, the quality of the generated LFs and the overall labeling accuracy improved substantially (Section~\ref{sec:sensitivity}). For instance, in the \textit{Finance} dataset, with 87 labeled examples in $D_l$, \tool generated 60 LFs (20 per category) to classify comments as \textit{positive}, \textit{neutral}, or \textit{negative}. The aggregated label for the comment \textit{``Thank you \$GOOG (Google Alphabet) and \$FB (Facebook) stocks! What a nice reversal.''} was predicted as \textit{neutral} because most weak labels abstained or voted \textit{neutral}, whereas the correct label is \textit{positive}. This illustrates how insufficient labeled supervision can lead to under-discriminative LFs and thus suboptimal aggregation.

\begin{figure}
    \centering
    \includegraphics[width=\linewidth]{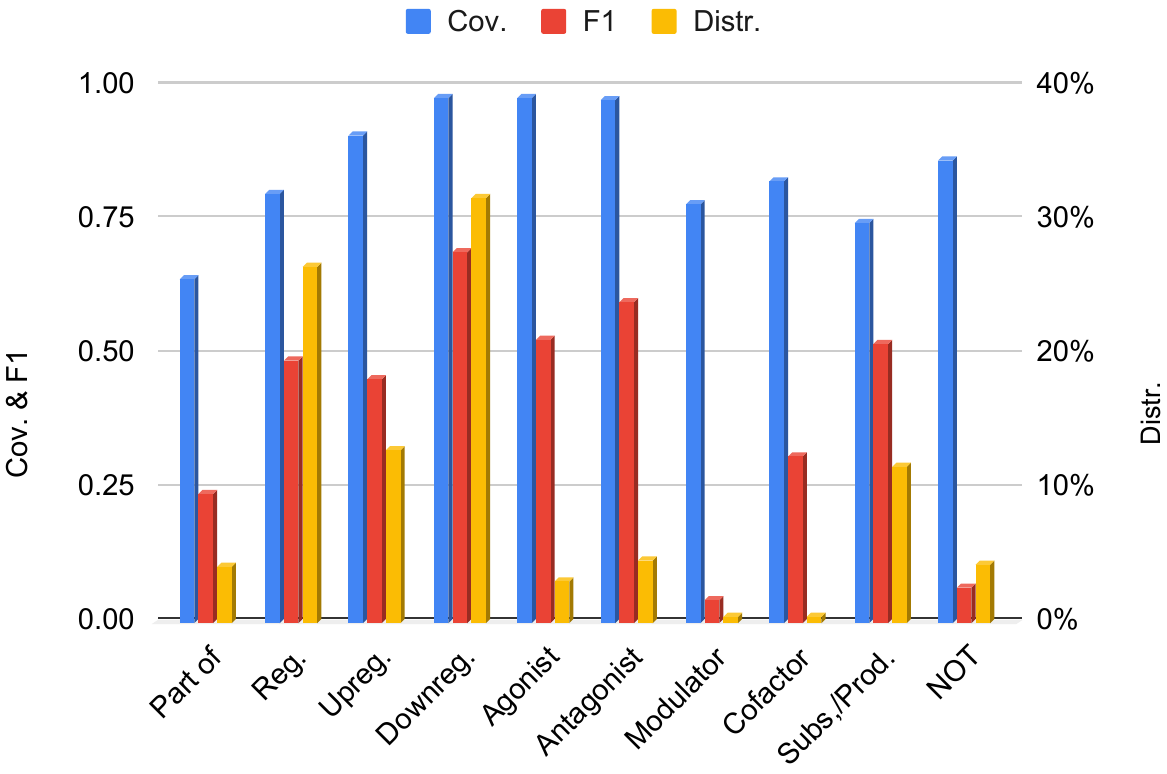}
    \caption{Per-class proportion of instances (orange bars) as well as per-class coverage and F1-Score of \tool for dataset \textit{ChemProt}.}
    \label{fig:chemprot}
\end{figure}

We also performed a class-wise analysis on a representative multi-class dataset, i.e., \textit{ChemProt}, reporting both coverage and per-class weighted F1-scores to examine \tool's behavior under severe class imbalance. {Figure\mbox{~\ref{fig:chemprot}} shows that \mbox{\tool} consistently maintained non-trivial coverage, including minority classes, although prediction quality degraded for extremely rare relations.}
This degradation is expected under extreme class imbalance, where only a handful of examples are available to induce reliable heuristics, suggesting that the limitation primarily stems from data scarcity rather than a failure of the proposed method.
In contrast, the other approaches often achieved zero F1 for the same rare classes (e.g., \datasculpt's F1-scores on \textit{Modulator}, \textit{Cofactor}, and \textit{Part of} are all zero) due to the absence of effective coverage, whereas \tool preserved non-zero coverage and measurable predictive performance.

We also observed several cases where other LF generation frameworks failed to produce meaningful coverage due to their inability to capture heterogeneous abstraction levels in text. For example, the Yelp review \textit{``Spoke to owner and they are not open to the public anymore. Info on website and Yelp etc. not up to date. Not sure why they have a storefront that says they are open 11-6.''} was not covered by any LF generated by either \alchemist or \datasculpt. 
An inspection of a representative LF produced by \alchemist revealed the limitation. Its generated rule relied exclusively on matching explicit lexical cues such as ``excellent'', ``friendly staff'', or ``worst'', as shown below\footnote{The full version of this LF could be found on our website~\cite{website}}:

\noindent\texttt{pos\_phrases = [``highly recommend'', ``amazing'', ``delicious'', ``great service'']}

\noindent\texttt{neg\_phrases = [``rude staff'', ``terrible'', ``overpriced'', ``bad experience'']}

\noindent Such heuristics can only recognize overt sentiment expressions and fail when the sentiment is implied through situational context, i.e., business closure or inconsistency between stated and actual operating hours. In this case, the review conveys dissatisfaction implicitly, without containing any of the surface-level sentiment keywords that \alchemist's rules depend on.
However, thanks to \tool's integration of structural and semantic LF generation, multiple LFs correctly identified this review as expressing a \textit{negative} sentiment. As a result, the aggregated label produced by \tool matched the ground truth. This showed that incorporating multi-level textual understanding, beyond surface heuristics, was essential for robust label generation in complex real-world datasets.

\subsection{Contribution of LF Categories}
\begin{table*}
\centering
\footnotesize
\caption{\tool's performance under controlled LF-category configurations, where only the specified LF category (or combination of categories) is enabled during LF exploration and inter-type filtering is disabled.}
\label{tab:lf-catetories}
\begin{tabular}{l|l|ccccc}\toprule
& \textbf{LF Config.} &\textbf{Coverage} &\textbf{Label. Quality} &\textbf{E2E Perf.} \\\midrule
\multirow{4}{*}{\textit{IMDb}} 
&Surface &0.953 &0.761 &0.739 \\
&Structural &0.998 &0.801 &0.740 \\
&Semantic &0.413 &0.345 &0.698 \\
&\textbf{\textit{Full config.}} (\tool) &1.000 &0.827 &0.748 \\\midrule
\multirow{4}{*}{\textit{Finance}} 
&Surface &1.000 &0.615 &0.576 \\
&Structural &0.956 &0.511 &0.514 \\
&Semantic &0.409 &0.288 &0.553 \\
&\textit{\textbf{Full config.}} (\tool) &1.000 &0.630 &0.612 \\\midrule
\multirow{4}{*}{\textit{Massive}} &Surface &0.920  &0.637 &0.555 \\
&Structural &0.770 &0.631 &0.595 \\
&Semantic &0.463 &0.346 &0.506 \\
&\textit{\textbf{Full config.}} (\tool) &0.976 &0.746 &0.598 \\
\bottomrule
\end{tabular}
\end{table*}


Table~\ref{tab:lf-catetories} shows the performance of \tool under different LF categories: \textit{Surface}, \textit{Structural}, and \textit{Semantic}.
Our results show that each LF category exhibited distinct characteristics. In particular, \textit{Surface LFs} generally achieved high coverage (close to or equal to 1.0 in Finance and IMDb), but their precision was limited, leading to moderate label quality and E2E performance. 
{\textit{Structural LFs} did not consistently improve label quality and E2E performance compared to surface heuristics, and coverage dropped substantially (e.g., in Massive).}
\textit{Semantic LFs} achieved the lowest label quality in isolation and suffered from extremely low coverage, which significantly reduced downstream effectiveness. 

However, combining LF categories in the full configuration of \tool consistently mitigated the weaknesses of individual categories. 
For example, in Finance, the tri-category setup yielded the best E2E performance, surpassing Semantic alone by +11\%. In IMDb, the combination achieved perfect coverage and the highest label quality, improving upon the strongest single category by 3\%--144\%. Overall, the results confirm that \tool benefited from the synergy of heterogeneous LF categories, with joint use delivering the most reliable labeling performance.

%
%

\subsection{Intrinsic Analysis}

\subsubsection{Impact of models used in generating Surface, Structural, and Semantic LFs}
\begin{table}
\centering
\footnotesize
\caption{Effect of the choice of LLM used to generate surface LFs on the performance of \tool. Dataset: \textit{IMDb}.}
\label{tab:llm-surface}
\footnotesize
\begin{tabular}{l|ccc}\toprule
\textbf{LLM} &\textbf{Coverage} &\textbf{Label. Quality} &\textbf{E2E Perf.} \\\midrule
GPT-5 &1.000 &0.828 &0.756 \\
GPT-4.1 &1.000 &0.827 &0.748 \\
Gemini-2.5-Flash &1.000 &0.814 &0.734 \\
Qwen-2.5-7B &0.999 &0.788 &0.732 \\
Gemma3-4B &1.000 &0.796 &0.717 \\
\bottomrule
\end{tabular}
\end{table}

We examined how the choice of LLM affects surface LF generation by building several \tool variants with different LLMs: \text{GPT-5}, \text{GPT-4.1}, \text{Gemini-2.5-Flash}, \text{Qwen-2.5-7B}, and \text{Gemma3-4B}, while keeping all other components fixed.
Results on the IMDb dataset (Table~\ref{tab:llm-surface}) show that stronger LLMs (\text{GPT-5} or \text{GPT-4.1}) yielded higher-quality LFs and better E2E performance. The \text{GPT-5}-based variant achieved the best results. Although all models achieved nearly perfect coverage, smaller open-source models (\text{Qwen-2.5-7B}, \text{Gemma3-4B}) showed clear degradation in labeling accuracy, indicating that the reasoning and linguistic capabilities of the underlying LLM strongly affect surface LF reliability.

\begin{table}
\centering
\footnotesize
\caption{Performance of \tool on \textit{Yelp} when different classifiers are used to generate Structural LFs.}
\label{tab:classifier-structural}
\footnotesize
\begin{tabular}{l|ccc}\toprule
\textbf{Classifier} &\textbf{Coverage} &\textbf{Label. Quality} &\textbf{E2E Perf.} \\\midrule
LR &0.990 &0.859 &0.850 \\
DT &0.978 &0.820 &0.847 \\
SVM &1.000 &0.865 &0.853 \\
KNN &0.983 &0.820 &0.845 \\
MLP &1.000 &0.857 &0.852 \\
\bottomrule
\end{tabular}
\end{table}

To study structural LFs, we compared lightweight classifiers, including Linear Regression (LR), Decision Tree (DT), Support Vector Machine (SVM), K-Nearest Neighbors (KNN), and Multi-Layer Perceptron (MLP), on the Yelp dataset.
As shown in Table~\ref{tab:classifier-structural}, SVM and MLP outperformed other classifiers, achieving the highest labeling quality and end-to-end F1. Specifically, the SVM variant achieved the best labeling quality (0.865) and the highest E2E performance (0.853). In contrast, DT and KNN showed lower quality, likely due to instability under sparse feature distributions. The coverage remained consistently high across all variants, suggesting SVM and MLP as robust choices for structural LF generation.

\begin{table}
\centering
\footnotesize
\caption{Performance of \tool on \textit{Massive} when different embedding models are used to generate Semantic LFs.}
\label{tab:pretrained-semantic}
\footnotesize
\begin{tabular}{l|ccc}\toprule
\textbf{Pretrained Model} &\textbf{Coverage} &\textbf{Label. Quality} &\textbf{E2E Perf.} \\\midrule
bert-base &0.976 &0.746 &0.598 \\
bge-base-en-v1.5 &1.000 &0.833 &0.636 \\
bart-base &0.992 &0.731 &0.605 \\
multilingual-e5-large &0.995 &0.817 &0.621 \\
\bottomrule
\end{tabular}
\end{table}

Finally, we evaluated semantic LFs using pretrained encoders \text{BERT-base}, \text{BGE-base-en-v1.5}, \text{BART-base}, and \text{Multilingual-E5} on the \text{Massive} dataset. 
Table~\ref{tab:pretrained-semantic} shows that embedding models with stronger semantic representations (\text{BGE-v1.5}, \text{Multilingual-E5}) produced more reliable LFs, outperforming earlier models such as \text{BERT} and \text{BART}. All variants maintained high coverage, suggesting that semantic LFs enhanced contextual consistency and complement surface and structural heuristics.

\subsubsection{Impact of Abstain Allowance in LFs}
\begin{table}\centering
\caption{Performance of \tool with and without abstain allowance in generating LFs.}
\label{tab:abstain}
\footnotesize
\begin{tabular}{l|l|ccc}\toprule
 & \textbf{Regime} &\textbf{Coverage} &\textbf{Label. Quality} &\textbf{E2E Perf.} \\\midrule
\multirow{2}{*}{\textit{YouTube}}
&Non abst. &1.000 &0.803 &0.782 \\
&Abstain &1.000 &0.848 &0.870 \\\midrule
\multirow{2}{*}{\textit{Agnews}} 
&Non abst. &1.000 &0.817 &0.796 \\
&Abstain &1.000 &0.855 &0.827 \\\midrule
\multirow{2}{*}{\textit{Finance}} 
&Non abst. &1.000 &0.599 &0.596 \\
&Abstain &1.000 &0.623 &0.617 \\
\bottomrule
\end{tabular}
\end{table}


Table~\ref{tab:abstain} compares \tool's performance when LFs were configured with and without abstain allowance. Specifically, the abstention setting allowed LFs to withhold predictions on uncertain instances, whereas the non-abstention setting did not.
Across all datasets, enabling abstention consistently improved both label quality and E2E performance while maintaining full coverage. Specifically, label quality increased by 4.5\%, 3.8\%, and 2.4\% absolute on \textit{YouTube}, \textit{Agnews}, and \textit{Finance}, respectively, leading to corresponding E2E performance gains of 8.8\%, 3.1\%, and 2.1\%. 
The consistent improvement suggests that allowing LFs to abstain on uncertain instances effectively reduces the propagation of noisy labels, enabling the label model to infer more accurate consensus labels and enhancing downstream learning. 
Notably, since all LFs preserved full coverage, the observed gains were derived purely from improved label correctness rather than selective filtering. These results highlight that abstention served as a reliable mechanism for noise control in weak supervision, particularly in heterogeneous or ambiguous labeling conditions. Consequently, abstention thresholds could be integrated in practice, allowing LFs to withhold low-confidence predictions, which in turn enhanced both intermediate label quality and overall model robustness.

\begin{figure}
    \centering
    \includegraphics[width=\linewidth]{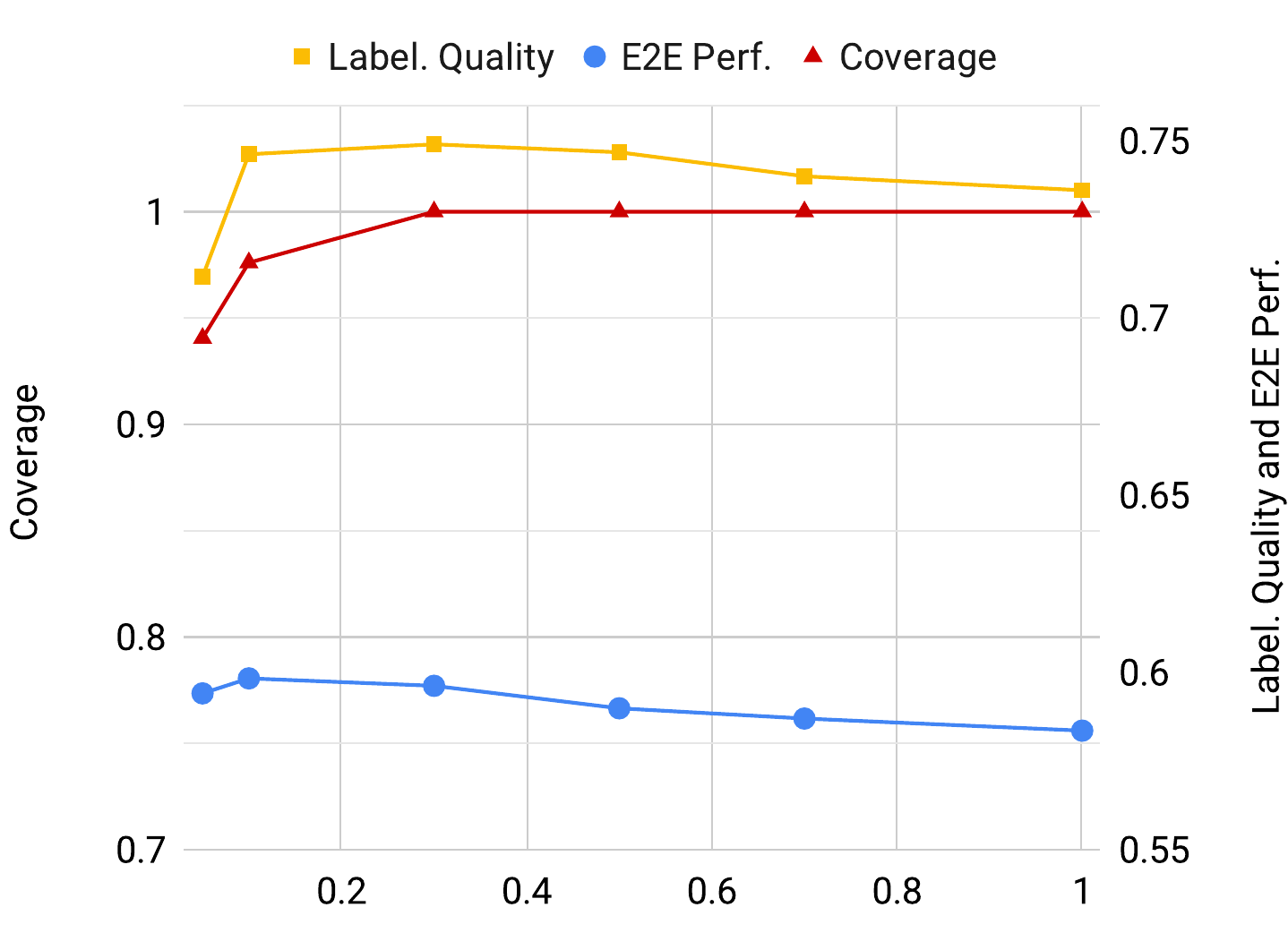}
    \caption{Impact of weighting factor $\beta$ when allowing abstain on \tool's performance. Dataset: \textit{Massive}.}
    \label{fig:beta-massive}
\end{figure}

Furthermore, we investigated the impact of the trade-off between precision and coverage when allowing abstention. Figure~\ref{fig:beta-massive} shows the performance of \tool with varied the weighting factor $\beta$. In our experiments, in \tool we configured $\beta < 1$, giving precision more weight. 
As shown in Figure~\ref{fig:beta-massive}, increasing $\beta$ from $0.05$ to $0.3$ improved label quality and end-to-end performance, indicating that a modest increase in the relative weight of coverage could benefit the calibration of the confidence thresholds $\omega_k$. However, as $\beta$ continued to increase beyond $0.3$, both label quality and E2E performance gradually declined, suggesting that excessive emphasis on coverage allows less-confident LFs to contribute noisier pseudo-labels. Coverage itself quickly saturated to nearly $1.0$ when $\beta \geq 0.1$, indicating that most LFs became sufficiently inclusive under moderate $\beta$ values. 
These results reinforce the importance of maintaining a precision-oriented calibration during threshold selection. A smaller $\beta$ (typically between $0.1$ and $0.3$) achieved the best trade-off, preserving high precision while maintaining nearly full coverage. This empirically supports our design choice of setting $\beta \ll 1$ to prioritize label reliability without sacrificing the diversity and completeness of weak supervision signals.

\subsubsection{Impact of Configuration of Label Function Exploitation}
\begin{figure}
    \centering
    \includegraphics[width=\linewidth]{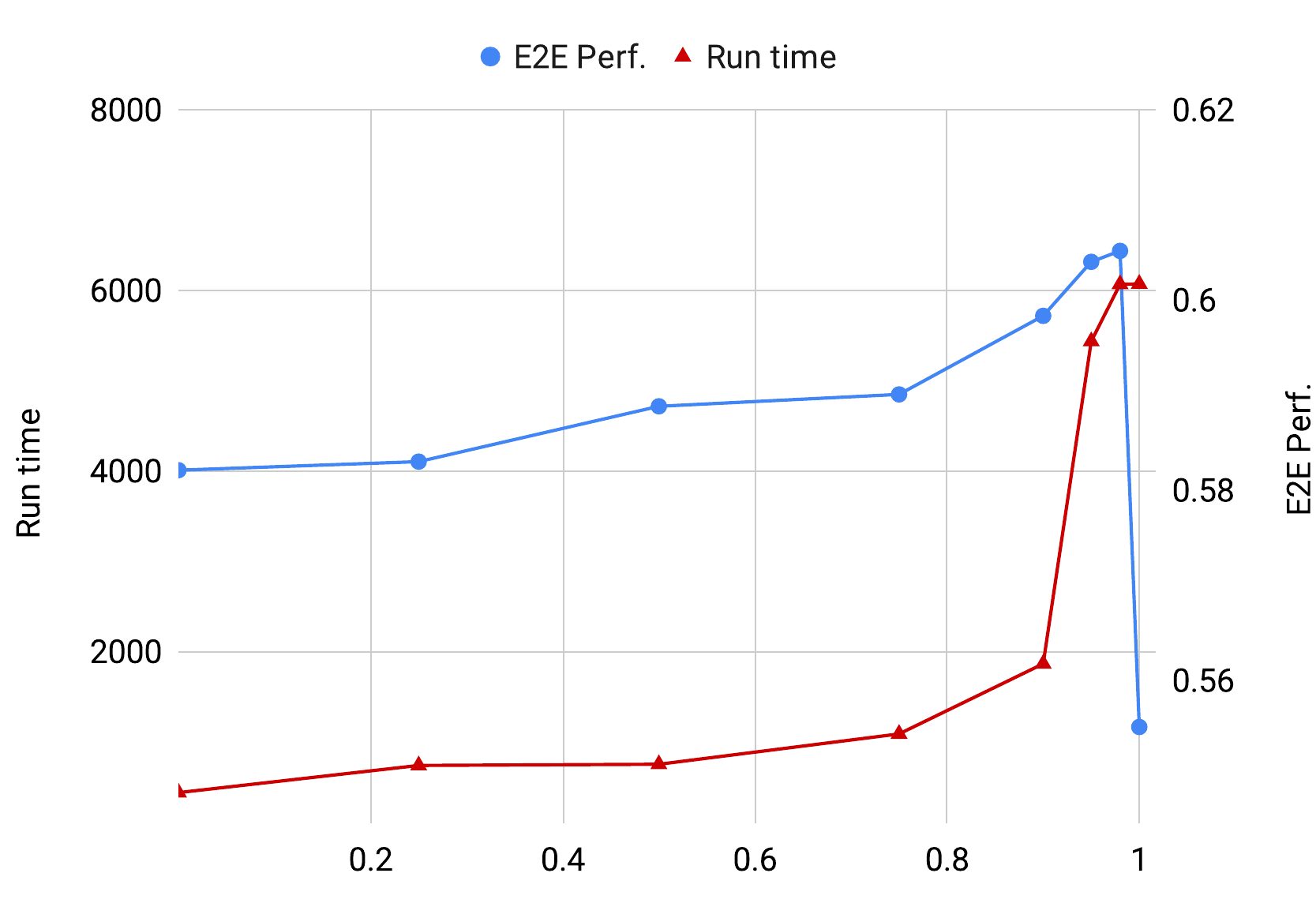}
    \caption{\tool's run time and E2E performance as a function of filtering parameter $\alpha$. Dataset: \textit{Massive}.}
    \label{fig:alpha}
\end{figure}

We further investigated how the configuration of the label function exploitation phase affects \tool's performance and efficiency by varying the acceptance multiplier $\alpha$. 
Recall that $\alpha$ controls the strictness of intra-type filtering, determining the minimum accuracy threshold $\theta^c_{\text{intra}} = \alpha \cdot \max_{\lambda_j \in \Lambda_c} \widehat{acc}(\lambda_j)$ within each LF category $c$. A smaller $\alpha$ permits a broader set of label functions, prioritizing diversity, while a larger $\alpha$ enforces stricter selection, retaining only the most reliable LFs. Setting $\alpha = 0$ disables exploitation, allowing all generated LFs to pass through unfiltered.

As shown in Figure~\ref{fig:alpha}, increasing $\alpha$ gradually improved E2E performance, with performance rising from 0.582 at $\alpha = 0$ to 0.605 at $\alpha = 0.98$. This trend shows that pruning noisy or inaccurate LFs yields a cleaner supervision signal, leading to higher-quality consensus labels and improved downstream model learning. However, this benefit came at the cost of higher computational overhead: the total runtime more than tripled from 437 seconds at $\alpha = 0$ to approximately 6K seconds at $\alpha = 0.98$. 
When $\alpha$ approached 1.0, performance began to saturate or even significantly decline, while the runtime continued to increase substantially. 
When $\alpha = 1$, generation was forcibly terminated to avoid an infinite loop.
Excessively high $\alpha$ values are likely to over-prune the LF pool (e.g., $\alpha = 1$ retained only the top-performing LFs in each category), thereby reducing diversity and limiting the complementary coverage across surface, structural, and semantic LFs. 
Overall, this trade-off shows that moderate $\alpha$ values (0.7--0.9) provide the best balance between label quality and efficiency, ensuring that the exploitation phase enhances LF reliability without compromising diversity or incurring excessive computational cost.

\subsubsection{Impact of LF size}

\begin{figure}
    \centering
    \includegraphics[width=\linewidth]{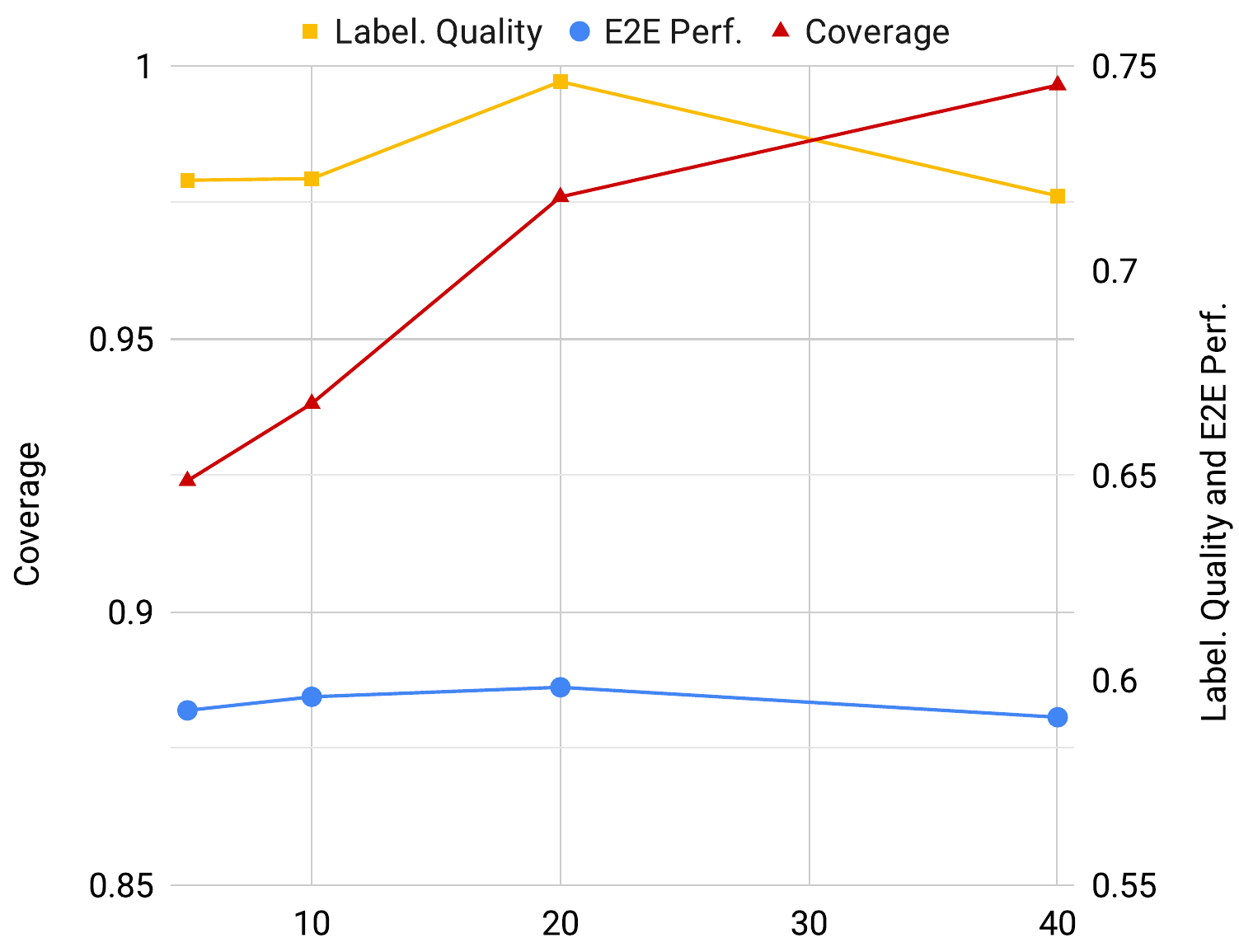}
    \caption{Coverage and performance of \tool as a function of the number of LFs per category $K_c$. Dataset: \textit{Massive}.}
    \label{fig:k-c}
\end{figure}

Beyond the exploitation configuration, we analyzed how the number of LFs per category influenced the \tool's performance. Figure~\ref{fig:k-c} shows the results in terms of label quality, coverage, and end-to-end performance on Massive.  

As seen, increasing the number of LFs per category ($K_c$) led to a clear rise in coverage, from $0.924$ with only 5 LFs to nearly full coverage ($0.997$) at 40 LFs. This trend suggests that larger LF pools increase the likelihood of covering diverse intents and linguistic variations in the data. However, the improvement in label quality and downstream E2E performance did not scale proportionally.

Label quality improved slightly as $K_c$ increases from 5 to 20, peaking at $0.746$ before declining when $K_c$ reached 40. The initial improvement reflects greater representational diversity among LFs, while the subsequent decline suggests that excessive LFs introduce redundancy and noise, reducing the reliability of aggregated labels. A similar trend was observed in E2E performance. This indicates that correlated or low-quality LFs can produce spurious consensus, ultimately harming downstream generalization.

{Overall, these results demonstrate that \textit{more LFs are not always better.}} While a larger LF set improved coverage, its marginal benefits on label quality and E2E performance quickly saturated. For complex datasets such as \textit{Massive}, an intermediate LF size (around 10--20 per category) provided the best trade-off between coverage and accuracy. We thus recommend an adaptive strategy that incrementally expands the LF pool until coverage stabilizes, ensuring both diversity and reliability in weak supervision.

\subsubsection{Impact of Label Model}
\begin{table}
\centering
\footnotesize
\caption{Influence of label models on the overall performance of \tool.}
\label{tab:label-model}
\footnotesize
\begin{tabular}{l|l|ccc}\toprule
\textbf{Dataset} &\textbf{Label Model} &\textbf{Label. Quality} &\textbf{E2E Perf.} \\\midrule
\multirow{5}{*}{\textit{IMDb}} 
&\mv &0.832 &0.756 \\
&\wmv &0.832 &0.757 \\
&\ds &0.826 &0.748 \\
&\fs &0.830 &0.755 \\
&\snorkel &0.829 &0.752 \\\midrule
\multirow{5}{*}{\textit{Finance}} 
&\mv &0.595 &0.561 \\
&\wmv &0.555 &0.535 \\
&\ds &0.641 &0.619 \\
&\fs &0.585 &0.501 \\
&\snorkel &0.638 &0.602 \\\midrule
\multirow{5}{*}{\textit{Massive}} 
&\mv &0.778 &0.618 \\
&\wmv &0.752 &0.602 \\
&\ds &0.751 &0.609 \\
&\fs &0.731 &0.490 \\
&\snorkel &0.773 &0.607 \\
\bottomrule
\end{tabular}
\end{table}


We further examined how different label models influence \tool's performance by comparing five representative aggregation strategies: Majority Vote, Weighted Majority Vote, Dawid-Skene, FlyingSquid, and Snorkel. As all models operated on the same set of LFs, their coverage remained constant, allowing us to focus on label quality and end-to-end performance.

In Table~\ref{tab:label-model}, the results show that \mv consistently provided a strong and stable baseline, achieving competitive label quality across datasets (e.g., $0.832$ on IMDb, $0.595$ on Finance, and $0.778$ on Massive). Despite its simplicity, the balanced aggregation of majority voting ensures robustness, leading to the highest or near-highest E2E scores in most settings. The performance of \wmv, which adjusts LF influence based on estimated reliability, varies across datasets. While marginally improving E2E performance in IMDb ($0.757$ vs. $0.756$ with \mv), it underperformed in Finance and Massive. More sophisticated probabilistic models such as \ds and \snorkel generally achieved higher label quality in challenging domains (e.g., \ds reached $0.641$ in Finance, surpassing \mv by nearly $8\%$ relative gain). Conversely, \fs, which assumes conditional independence among LFs, produced less stable outcomes, particularly on datasets with correlated or overlapping LFs (e.g., E2E drops to $0.490$ on Massive).

Overall, the results show \tool's flexibility in accommodating a broad range of label models. Regardless of the underlying aggregation mechanism, \tool consistently translates modest gains in label quality into proportionate improvements in E2E performance. These findings suggest that while advanced models can provide incremental benefits in complex labeling scenarios, even simple approaches such as \mv remain remarkably effective when integrated within \tool's pipeline.
\subsection{Sensitivity Analysis}
\label{sec:sensitivity}

\begin{figure}
    \centering
    \includegraphics[width=\linewidth]{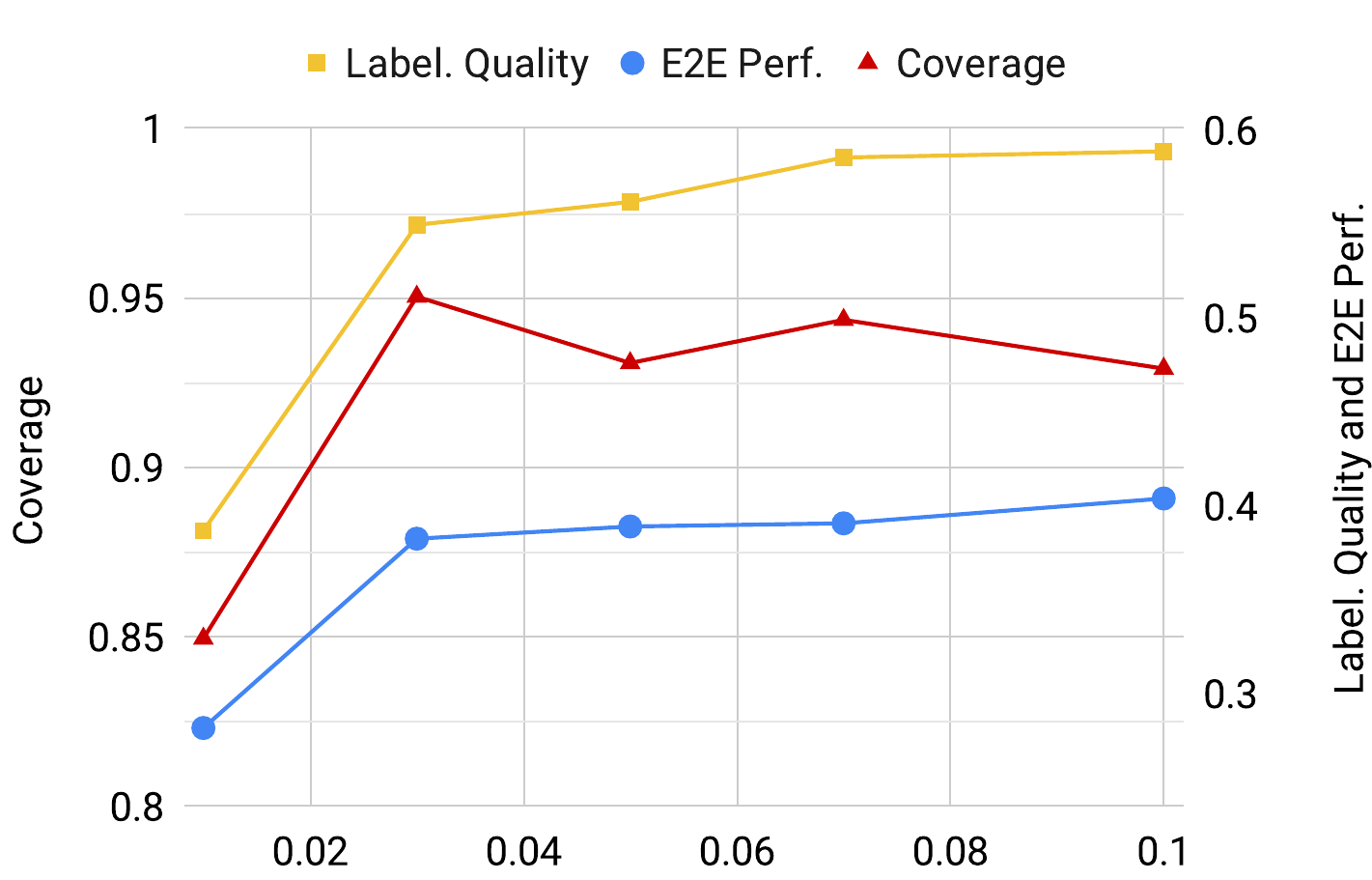}
    \caption{Coverage and performance of \tool as a function of the proportion of labeled instances ($|D_l|/(|D|+|D_l|)$). Dataset: \textit{Chemprot}.}
    \label{fig:D-l}
\end{figure}

Figure~\ref{fig:D-l} shows the effect of varying the proportion of labeled data on \tool's performance for \textit{Chemprot}. 
Overall, \tool exhibited a consistent upward trend across all metrics as the fraction of labeled data increased, confirming its capacity to effectively utilize additional supervision. 

At very low supervision levels (e.g., 1\% labeled data), both label quality and E2E performance remained modest, reflecting the inherent difficulty of the biomedical relation extraction task under sparse annotation. 
However, a notable improvement was observed when the labeled proportion increased to 3\%, indicating that even a small amount of additional supervision can significantly enhance both label quality and downstream task performance. 
Beyond this point, the improvement became more incremental, suggesting that \tool rapidly learned from limited supervision and progressively converges as label availability grows. 
These findings highlight \tool's robustness in low-resource settings and its ability to deliver competitive performance with minimal labeled data, while continuing to benefit, albeit with diminishing returns, from further supervision.
\subsection{Efficiency Analysis}

All experiments were conducted on a Linux 5.15.154 server equipped with two NVIDIA T4 GPUs. 
The average labeling time ranged from 0.08 to 0.28 seconds per example, depending on dataset size and task complexity. 
Within the programmatic labeling pipeline, LF generation (exploration and exploitation) accounted for 47\%--86\% of the total labeling time, depending on dataset size, with LF exploration ($K_c = 20$) identified as the most time-consuming phase, contributing roughly 99\% of the LF generation process. 
A deeper analysis of the LF exploration phase revealed that the distribution of exploration time across different LF categories was 42.26\% for surface-based LFs (generated via GPT-4.1), 55.07\% for structural LFs (using TF-IDF with an SVM classifier), and 2.67\% for semantic LFs (using a BERT encoder with an MLP containing one hidden layer). 
In terms of cost efficiency, the labeling cost for 1,000 examples using GPT-4.1 ranged from \$0.005 to \$0.130. These results demonstrate that \tool achieves high-quality programmatic labeling at an inexpensive computational cost.

\subsection{Practical Considerations}

{Beyond the controlled experimental setting, we discuss several practical considerations for deploying \mbox{tool} in real-world classification scenarios.}

As observed in Table~\ref{tab:single_label}, \tool, like other programmatic labeling frameworks, exhibits reduced performance on large multi-class datasets such as \textit{ChemProt} and \textit{Massive}. This degradation was primarily attributable to increased class ambiguity and severe class imbalance, which amplify the difficulty of designing high-precision label functions that generalize across many fine-grained classes. Nevertheless, the use of heterogeneous LF categories (surface, structural, and semantic) enables \tool to capture complementary signals, partially mitigating this challenge. These results suggest that while \tool was effective in multi-class settings, further improvements may be achieved by incorporating class-aware LF generation or adaptive thresholding strategies for underrepresented classes.

{In scenarios where class definitions or topical distributions evolve, \mbox{\tool} is well positioned to support efficient re-labeling with minimal manual supervision.} Since \tool relies on a small seed set of labeled data and programmatically generated LFs, it can be periodically re-invoked to regenerate or update LFs as new topics emerge or existing ones shift. In practice, this allows practitioners to incrementally refresh the weak supervision pipeline by (i) updating the seed set with a small number of newly labeled instances, (ii) re-running the LF exploration phase, and (iii) re-applying the exploitation phase to filter outdated or noisy LFs. This lightweight adaptation mechanism makes \tool particularly suitable for dynamic environments where continuous full re-annotation is infeasible.

\subsection{Threats to Validity}

The main threats to the validity of our work consist of internal, external, and construct validity threats.

\textbf{Threats to internal validity:}
A potential internal threat concerns the correctness of our implementation, the design of the automated LF generation process, and the configuration of hyperparameters for LF selection and aggregation. To minimize this risk, we thoroughly reviewed our codebase, validated key components of the LF exploration and exploitation phases, and performed extensive ablation studies to ensure the robustness of our design choices. All hyperparameters were tuned systematically through multiple trials. Furthermore, we have made our implementation publicly available~\cite{website} to facilitate independent verification and reproducibility.

\textbf{Threats to external validity:}
External threats are primarily related to the generalizability of our findings. Although we evaluated \tool across eleven publicly available text classification benchmarks spanning diverse domains (e.g., sentiment analysis, spam detection, biomedical relation extraction), our results may not fully generalize to other modalities such as images, audio, or multi-label tasks. In addition, while these datasets are widely used in the weak supervision literature, they may not capture all forms of noise and task complexity encountered in real-world annotation pipelines. To reduce this threat, we selected datasets with varying label distributions, class granularities, and difficulty levels. Future work will explore applying \tool to multimodal and industrial-scale datasets to assess generalizability further. 
%

\textbf{Threats to construct validity:}
Construct validity threats may arise from the choice of evaluation metrics and baselines. To ensure fairness, we adopted standard weak supervision metrics (coverage, label accuracy, and weighted F1 for downstream models) that are widely used in prior studies~\cite{huang2024alchemist, guan2025datasculpt}. Another potential threat lies in the use of specific LLMs and aggregation models that may influence performance. To address this, we conducted controlled experiments with consistent setups across all baselines. Nevertheless, results may vary when different foundation models or aggregation strategies are employed. Future extensions could incorporate adaptive model selection to further validate our framework under different conditions.

\section{Related Work}
\label{sec:related_work}

\tool closely relates to recent studies that explored the \textbf{automation of LF generation}, aiming to alleviate the manual effort in programmatic labeling.
%
In particular, Alchemist~\cite{huang2024alchemist} and DataSculpt~\cite{guan2025datasculpt} leverage LLMs to generate labeling programs or rules automatically. Alchemist employs multiple prompting regimes, optionally integrating retrieval-augmented generation (RAG), to derive surface-level heuristics from task descriptions. Similarly, DataSculpt prompts LLMs to produce keyword-, pattern-, or statistical-based LFs. 
However, both rely heavily on LLMs' textual reasoning to synthesize \textit{surface} heuristics, which often lack \textit{structural} and \textit{semantic} dimensions crucial for complex tasks. In contrast, \tool formulates a principled framework to explore complementary LF categories, ensuring systematic coverage of diverse signal spaces rather than relying on direct prompt outputs. While Alchemist and DataSculpt primarily focus on generation quality through prompt design, their methods could be integrated into \tool's surface-level component to enrich LF diversity.

Snuba~\cite{varma2018snuba} is a model-based heuristic synthesis that automates weak supervision by generating simple heuristic models (e.g., decision trees) from a small labeled dataset and user-defined primitives. It iteratively balances accuracy and coverage, abstains when uncertain, and terminates automatically when further heuristics risk degrading label quality. 
While Snuba focuses on generating lightweight classifiers as LFs, \tool systematically explores diverse categories of LFs, from \textit{surface}, \textit{structural}, to \textit{semantic}, to better capture multi-level data signals. As Snuba relies solely on feature-based decision boundaries, it often requires numerous heuristics yet still struggles to achieve high label quality. \tool unifies heterogeneous LF types in a structured manner, resulting in a more compact and higher-quality labeling set.

GLaRA~\cite{zhao2021glara} builds a graph of candidate rules extracted from unlabeled data, leveraging semantic similarity propagation via a GNN to augment labeling confidence. TALLOR~\cite{li2021weakly} incrementally composes logical rules from seed rules and pseudo-labeled data to form more expressive compound heuristics. 
Both methods emphasize structure-aware rule refinement and augmentation. Meanwhile, \tool focuses on exploring independent heuristics across levels rather than composing or propagating from seeds. Yet, the graph-based reasoning in GLaRA and the compositional logic of TALLOR could complement \tool, e.g., by promoting more semantically consistent or composite LFs.

IWS~\cite{boecking2020interactive} and DARWIN~\cite{galhotra2021adaptive} introduce interactive mechanisms where the system proposes candidate heuristics and experts provide meta-feedback on rule quality rather than labeling individual data points. This reduces annotation cost and shifts human effort toward reviewing high-level concepts. 
While \tool is primarily automated, integrating such feedback-driven refinement could further enhance the interpretability and reliability of generated LFs, especially in domain-specific contexts where expert intuition helps validate or prune weak signals.

\textbf{Programmatic labeling}~\cite{data-programming, fu2020fast, goggles, interactive-ws, ratner2017snorkel, ruan2025, oliveira2025, zhang2024stronger} provides a foundational paradigm for generating training labels by combining outputs from multiple noisy heuristics. Early work on data programming~\cite{data-programming, ratner2017snorkel} introduced the notion of label functions and proposed probabilistic label models to estimate true labels from conflicting weak signals. Later frameworks, such as GOGGLES~\cite{goggles} and FLYINGSQUID~\cite{fu2020fast}, extended this idea to vision and structured domains, while interactive weak supervision~\cite{interactive-ws} incorporated user feedback to refine heuristics. Despite their success in formalizing weak supervision, these approaches presuppose the existence of a high-quality set of LFs and primarily focus on label aggregation rather than LF generation. Meanwhile, \tool addresses the complementary challenge of \textit{automatically discovering and calibrating} LFs across different abstraction levels (surface, structural, semantic), ensuring both LF diversity and reliability before aggregation. As such, \tool can serve as a front-end LF synthesis module that seamlessly integrates with existing weak supervision frameworks to produce higher-quality labeling sources for downstream probabilistic models.

\tool is also related to latter advances in \textbf{automated data annotation}, which aim to reduce human labeling effort by leveraging LLMs, active learning, and semi-supervised strategies.
Recent work has leveraged LLMs as annotators~\cite{gpt-3-labeling, annollm, LLM-few-short, zerogen, tuning-lm, llmaaa, schroeder2025}, enabling zero- or few-shot labeling across diverse domains. While such models can directly assign labels with minimal supervision, their predictions often lack task-specific reliability and interpretability, particularly when label semantics are subtle or domain-dependent.
In parallel, active learning approaches~\cite{active-learning-book, active-learning-survey, active-learning-survey-2, guan2024activedp, al-hybridization, tbal} focus on selecting the most informative samples for manual annotation, while threshold-based automated labeling~\cite{tbal, vishwakarma2024pearls} relies on model confidence to propagate labels.
Classical semi-supervised learning~\cite{wang2006label, semi, zhou2003learning} also exploits structure and smoothness assumptions to infer missing labels from a small labeled subset.

Recently, several approaches have been proposed to address the challenges in \textbf{improving the data quality and quantity}, and eventually ML models' performance, such as corrupted label detection/cleaning in noisy data~\cite{cola, simifeat,data-fault-localization, kim2025delving}, data reduction~\cite{d2-pruning}, data synthesis/augmentation~\cite{data-synthetic}, outlier detections~\cite{outlier-detection}, and missing value imputation~\cite{missing-value}.

\section{Conclusion}
\label{sec:conclusion}

This work introduces \tool, an automated framework for programmatic labeling that systematically generates, selects, and calibrates LFs to enable reliable weak supervision. Unlike prior approaches that depend solely on LLM-generated heuristics or static rule synthesis, \tool formulates LF construction as a structured exploration and exploitation process. By explicitly balancing LF \textit{diversity} and \textit{reliability} during generation and selection, \tool produces a compact yet effective set of LFs that improves the overall quality of weakly labeled data.

Extensive experiments across diverse text classification tasks demonstrated that \tool consistently achieves superior labeling coverage and accuracy compared to the existing approaches.
These improvements lead to substantial gains in downstream model performance, reaching up to 46\% improvement in weighted F1. These results demonstrate that integrating multi-level LFs spanning surface, structural, and semantic perspectives within a unified framework enables \tool to produce more reliable weak labels and achieve stronger downstream performance.

Beyond empirical improvements, this work provides a broader insight into automated data annotation. Scalable weak supervision requires not only increasing the number of heuristics but also ensuring their reliability through systematic filtering and calibration.
This highlights the importance of principled LF selection and calibration in reducing noise and redundancy in programmatic labeling pipelines.

Despite its strong performance, \tool has limitations. In highly complex multi-class settings such as \textit{ChemProt} and \textit{Massive}, performance gains remain to be improved, indicating that additional task-specific signals or richer supervision may be required. Moreover, while \tool is robust under limited labeled data, its effectiveness still depends on the representativeness of the initial labeled subset.

Future work includes extending \tool to dynamically evolving data streams, incorporating adaptive LF retirement and regeneration mechanisms, and exploring tighter integration with active learning. Additionally, we plan to extend \tool to support multimodal data and interactive human-in-the-loop refinement, advancing toward fully automated yet verifiable data annotation. We believe this work contributes a practical step toward scalable, data-centric AI development by bridging heuristic synthesis, weak supervision, and LLM reasoning within a unified framework.

\printcredits
\bibliographystyle{elsarticle-num}

\bibliography{ref}

\end{document}